\documentclass{article}

\usepackage[final, nonatbib]{neurips_2020} 

\usepackage[utf8]{inputenc} 
\usepackage[T1]{fontenc}    
\usepackage{hyperref}       
\usepackage{url}            
\usepackage{booktabs}       
\usepackage{amsfonts}       
\usepackage{nicefrac}       
\usepackage{microtype}      
\usepackage{amsthm}
\usepackage{graphicx}      
\usepackage{subcaption}
\usepackage{bbm} 
\usepackage{amsmath}

\DeclareMathOperator*{\argmin}{arg\,min}
\usepackage{mathtools} 

\usepackage{algorithm}      
\usepackage[noend]{algpseudocode}

\usepackage{xcolor}

\newcommand{\algnamenospace}{BanditPAM}
\newcommand{\algname}{BanditPAM }

\newtheorem{theorem}{Theorem}

\newcommand{\Star}{\mathcal{S}_{\text{tar}}}

\newcommand{\nuref}{n_{\text{used\_ref}}}

\newcommand{\X}{{\mathcal X}}

\newcommand{\aln}[1]{\begin{align*}#1\end{align*}}

\newcommand{\al}[1]{\begin{align}#1\end{align}}

\newcounter{numcount}
\newif\iflong
\longfalse

\newif\ifdraft
\drafttrue

\renewcommand{\paragraph}[1]{\noindent {\bf #1}}

\title{\algnamenospace: Almost Linear Time $k$-Medoids Clustering via Multi-Armed Bandits}

\author{%
  Mo Tiwari  \\
  Department of Computer Science\\
  Stanford University\\
  \texttt{motiwari@stanford.edu} \\
  \And
  Martin Jinye Zhang \\
  Department of Epidemiology\\
  Harvard T.H. Chan School of Public Health\\
  \texttt{jinyezhang@hsph.harvard.edu} \\
  \And
  James Mayclin \\
  Department of Computer Science\\
  Stanford University\\
  \texttt{jmayclin@stanford.edu} \\
  \And
  Sebastian Thrun \\
  Department of Computer Science\\
  Stanford University\\
  \texttt{thrun@stanford.edu} \\
  \And
  Chris Piech \\
  Department of Computer Science\\
  Stanford University\\
  \texttt{piech@cs.stanford.edu} \\
  \And
  Ilan Shomorony\\
  Electrical and Computer Engineering\\
  University of Illinois at Urbana-Champaign\\
  \texttt{ilans@illinois.edu}
}

\begin{document}

\maketitle

\begin{abstract}


Clustering is a ubiquitous task in data science.
Compared to the commonly used $k$-means clustering, $k$-medoids clustering requires the cluster centers to be actual data points and supports arbitrary distance metrics, which permits greater interpretability and the clustering of structured objects.
Current state-of-the-art $k$-medoids clustering algorithms, such as Partitioning Around Medoids (PAM), are iterative and are quadratic in the dataset size $n$ for each iteration, being prohibitively expensive for large datasets.
We propose \algnamenospace, a randomized algorithm inspired by techniques from multi-armed bandits, that reduces the complexity of each PAM iteration from $O(n^2)$ to $O(n\log n)$ and returns the same results with high probability, under assumptions on the data that often hold in practice.
As such, \algname matches state-of-the-art clustering loss while reaching solutions much faster.
We empirically validate our results on several large real-world datasets, including a coding exercise submissions dataset from Code.org, the 10x Genomics 68k PBMC single-cell RNA sequencing dataset, and the MNIST handwritten digits dataset.
In these experiments, we observe that \algname returns the same results as state-of-the-art PAM-like algorithms up to 4x faster while performing up to 200x fewer distance computations. 
The improvements demonstrated by \algname enable $k$-medoids clustering on a wide range of applications, including identifying cell types in large-scale single-cell data and providing scalable feedback for students learning computer science online. We also release highly optimized Python and C++ implementations of our algorithm\footnote{https://github.com/ThrunGroup/BanditPAM}.

\end{abstract}


\section{Introduction \label{sec:intro}}

Many modern data science applications require the clustering of very-large-scale data. 
Due to its computational efficiency, the $k$-means clustering algorithm \cite{macqueen1967some,lloyd1982least} is one of the most widely-used clustering algorithms.
$k$-means alternates between assigning points to their nearest cluster centers and recomputing those centers. 
Central to its success is the specific choice of the cluster center: for a set of points, $k$-means defines the cluster center as the point with the smallest average \emph{squared Euclidean distance} to all other points in the set. 
Under such a definition, the cluster center is the arithmetic mean of the cluster's points and can be computed efficiently. 

While commonly used in practice, $k$-means clustering suffers from several drawbacks. 
Firstly, while one can efficiently compute the cluster centers under squared Euclidean distance, it is not straightforward to generalize to other distance metrics \cite{overton1983quadratically,jain1988algorithms,bradley1997clustering}. 
However, different distance  metrics may be desirable in other applications.
For example, $l_1$ and cosine distance are often used in sparse data, such as in recommendation systems \cite{leskovec2020mining} and single-cell RNA-seq analysis \cite{ntranos2016fast}; additional examples include string edit distance in text data \cite{navarro2001guided}, and graph metrics in social network data \cite{mishra2007clustering}. 
Secondly, the cluster center in $k$-means clustering is in general not a point in the dataset and may not be interpretable in many applications. This is especially problematic when the data is structured, such as parse trees in context-free grammars, sparse data in recommendation systems \cite{leskovec2020mining}, or images in computer vision where the mean image is visually random noise \cite{leskovec2020mining}. 

Alternatively, $k$-medoids clustering algorithms \cite{kaufman1987clustering,kaufman1990partitioning} use \emph{medoids} to define the cluster center for a set of points, where for a set and an arbitrary distance function, the medoid is the point \emph{in the set} that minimizes the average distance to all the other points.
Mathematically, for $n$ data points $\mathcal{X} =  \{ x_1, \cdots, x_n \}$ and a user-specified distance function $d(\cdot, \cdot)$, the $k$-medoids problem is to find a set of $k$ medoids $\mathcal{M} = \{m_1, \cdots, m_k \} \subset \mathcal{X}$ to minimize the overall distance of points to their closest medoids: 
\begin{equation} \label{eqn:sec_intro_total-loss}
    L(\mathcal{M}) =  \sum_{i=1}^n \min_{m \in \mathcal{M}} d(m, x_i)
\end{equation}
Note that the distance function can be arbitrary; indeed, it need not be a distance metric at all and could be an asymmetric dissimilarity measure. The ability to use an arbitrary dissimilarity measure with the $k$-medoids algorithm addresses the first shortcoming of $k$-means discussed above.
Moreover, unlike $k$-means, the cluster centers in $k$-medoids (i.e., the medoids) must be points in the dataset, thus addressing the second shortcoming of $k$-means clustering described above.


Despite its advantages, $k$-medoids clustering is less popular than $k$-means due to its computational cost. Problem \ref{eqn:sec_intro_total-loss} is NP-hard in general \cite{schubert2019faster}, although heuristic solutions exist.
Current state-of-the-art heuristic $k$-medoids algorithms scale quadratically in the dataset size in each iteration. However, they are still significantly slower than $k$-means, which scales linearly in dataset size in each iteration. 

Partitioning Around Medoids (PAM) \cite{kaufman1987clustering, kaufman1990partitioning} is one of the most widely used heuristic algorithms for $k$-medoids clustering, largely because it produces the best clustering quality \cite{reynolds2006clustering,schubert2019faster}.
PAM is split into two subroutines: BUILD and SWAP.
First, in the BUILD step, PAM aims to find an initial set of $k$ medoids by greedily and iteratively selecting points that minimize the $k$-medoids clustering loss (Equation \eqref{eqn:sec_intro_total-loss}). 
Next, in the SWAP step, PAM considers all $k(n-k)$ possible pairs of medoid and non-medoid points and swaps the pair that reduces the loss the most. 
The SWAP step is repeated until no further improvements can be made by swapping medoids with non-medoids. 
As noted above, PAM has been empirically shown to produce better results than other popular $k$-medoids clustering algorithms.
However, the BUILD step and each of the SWAP steps require  $O(kn^2)$ distance evaluations and can be prohibitively expensive to run, especially for large datasets or when the distance evaluations are themselves expensive (e.g. for edit distance between long strings).

Randomized algorithms like CLARA \cite{kaufman1990partitioning} and CLARANS \cite{ng2002clarans} have been proposed to improve computational efficiency, but result in worse clustering quality.
More recently, Schubert et al. \cite{schubert2019faster} proposed a deterministic algorithm, dubbed FastPAM1, that guarantees the same output as PAM but improves the complexity to $O(n^2)$.
However, the factor $O(k)$ improvement becomes less important when the sample size $n$ is large and the number of medoids $k$ is small compared to $n$.  Throughout the rest of this work, we treat $k$ fixed and assume $k \ll n$.

\paragraph{Additional related work:} Many other $k$-medoids algorithms exist. These algorithms can generally be divided into those that agree with or produce comparable results to PAM (matching state-of-the-art clustering quality, such as FastPAM and FastPAM1 \cite{schubert2019elki}) and other randomized algorithms that sacrifice clustering quality for runtime (such as CLARA and CLARANS). Park et al.~\cite{park2009simple} proposed a $k$-means-like algorithm that alternates between reassigning the points to their closest medoid and recomputing the medoid for each cluster until the $k$-medoids clustering loss can no longer be improved. 
Other proposals include optimizations for Euclidean space and tabu search heuristics \cite{estivill2001robust}. Recent work has also focused on distributed PAM, where the dataset cannot fit on one machine \cite{song2017pamae}. All of these algorithms, however, scale quadratically in dataset size or concede the final clustering quality for improvements in runtime. In an alternate approach for the single medoid problem, trimed \cite{trimed}, scales sub-quadratically in dataset size but exponentially in the dimensionality of the points. Other recent work \cite{activekmedoids} attempts to minimize the number of \textit{unique} pairwise distances required. Similarly, \cite{heckel} attempts to adaptively estimate these distances in specific settings.

\paragraph{Contributions:} 
In this work, we propose a novel randomized $k$-medoids algorithm, called \algnamenospace, that runs significantly faster than state-of-the-art $k$-medoids algorithms and achieves the same clustering results with high probability.
Modeled after PAM, \algname reduces the complexity on the sample size $n$ from $O(n^2)$ to $O(n\log n)$, for the BUILD step and each SWAP step, under reasonable assumptions that hold in many practical datasets.
We empirically validate our results on several large, real-world datasets
and observe that \algname provides a reduction of distance evaluations of up to 200x while returning the same results as PAM and FastPAM1.
We also release a high-performance C++ implementation of \algnamenospace, callable from Python, which runs 4x faster than the state-of-the-art FastPAM1 implementation on the full MNIST dataset ($n = 70,000$) -- without precomputing and caching the $O(n^2)$ pairwise distances as in FastPAM1.

Intuitively, \algname works by recasting each step of PAM from a \emph{deterministic computational problem} to a \emph{statistical estimation problem}. 
In the BUILD step assignment of the $l$th medoid, for example, we need to choose the point amongst all $n-l$ non-medoids that will lead to the lowest overall loss (Equation \eqref{eqn:sec_intro_total-loss}) if chosen as the next medoid. Thus, we wish to find $x$ that minimizes
\begin{equation} \label{eqn:sec_intro_build_loss}
	L(\mathcal{M}; x) = \sum_{j=1}^n \min_{m \in \mathcal{M} \cup \{x\}} d(m, x_j) =\vcentcolon \sum_{j=1}^n g(x_j),
\end{equation}
where $g(\cdot)$ is a function that depends on $\mathcal{M}$ and $x$.
Eq.~\eqref{eqn:sec_intro_build_loss} shows that the loss of a new medoid assignment $L(\mathcal{M}; x)$ can be written as the summation of the value of the function $g(\cdot)$ evaluated on all $n$ points in the dataset. Though approaches such as PAM and FastPAM1 compute $L(\mathcal{M}; x)$ exactly for each $x$, \algname \textit{adaptively estimates} this quantity by sampling reference points $x_j$ for the most promising candidates. Indeed, computing $L(\mathcal{M}; x)$ exactly for every $x$ is not required; promising candidates can be estimated with higher accuracy (by computing $g$ on more reference points $x_j$) and less promising ones can be discarded early without expending further computation.

To design the adaptive sampling strategy, we show that the BUILD step and each SWAP iteration can be formulated as a best-arm identification problem from the multi-armed bandits (MAB) literature \cite{audibert2010best,even2002pac,jamieson2014lil,jamieson2014best}. 
In the typical version of the best-arm identification problem, we have $m$ arms. At each time step $t = 0,1,...,$ we decide to pull an arm $A_t\in \{1,\cdots,m\}$, and receive a reward $R_t$ with $E[R_t] = \mu_{A_t}$. The goal is to identify the arm with the largest expected reward with high probability with the fewest number of total arm pulls.
In the BUILD step of \algnamenospace, we view each candidate medoid $x$ as an arm in a best-arm identification problem. The arm parameter corresponds to $\tfrac{1}{n}\sum_j g(x_j)$ 
and pulling an arm corresponds to computing the loss $g$ on a randomly sampled data point $x_j$. Using this reduction, the best candidate medoid can be estimated using existing best-arm algorithms like the Upper Confidence Bound (UCB) algorithm \cite{lai1985asymptotically} and successive elimination \cite{successiveelimination}.



The idea of algorithm acceleration by converting a computational problem into a statistical estimation problem and designing the adaptive sampling procedure via multi-armed bandits has
witnessed some recent successes \cite{chang2005adaptive,kocsis2006bandit,10.5555/3122009.3242042,jamieson2016non,bagaria2018adaptive,zhang2019adaptive}.
In the context of $k$-medoids clustering, previous work \cite{bagaria2018medoids,baharav2019ultra} has considered finding the \textit{single} medoid of a set points (i.e., the $1$-medoid problem).
In these works, the $1$-medoid problem was also formulated as a best-arm identification problem, with each point corresponding to an arm and its average distance to other points corresponding to the arm parameter. 

While the $1$-medoid problem considered in prior work can be solved exactly, the $k$-medoids problem is NP-Hard and is therefore only tractable with heuristic solutions. Hence, this paper focuses on improving the computational efficiency of an existing heuristic solution, PAM, that has been empirically observed to be superior to other techniques.
Moreover, instead of having a single best-arm identification problem as in the $1$-medoid problem, we reformulate PAM as a \textit{sequence} of best-arm problems. Our reformulation treats different objects as arms in different steps of PAM; in the BUILD step, each point corresponds to an arm, whereas in the SWAP step, each medoid-and-non-medoid pair corresponds to an arm.
We notice that the intrinsic difficulties of this sequence of best-arm problems are different from the single best-arm identification problem, which can be exploited to further speed up the algorithm. We discuss these further optimizations in Sections \ref{sec:exps} and \ref{sec:discussion} and Appendix \ref{A1}.



\section{Preliminaries \label{sec:prelims}}
For $n$ data points $\mathcal{X} =  \{ x_1, x_2, \cdots, x_n \}$ and a user-specified distance function $d(\cdot, \cdot)$, the $k$-medoids problem aims to find a set of $k$ medoids $\mathcal{M} = \{m_1, \cdots, m_k \} \subset \mathcal{X}$ to minimize the overall distance of points from their closest medoids:
\begin{equation} \label{eqn:total-loss}
	L(\mathcal{M}) =  \sum_{i=1}^n \min_{m \in \mathcal{M}} d(m, x_i)
\end{equation}
Note that $d$ need not satisfy symmetry, triangle inequality, or positivity. 
For the rest of the paper, we use $[n]$ to denote the set $\{1,\cdots,n\}$ and $\vert \mathcal{S} \vert$ to represent the cardinality of a set $\mathcal{S}$.
For two scalars $a,b$, we let $a\wedge b = \min(a,b)$ and $a\vee b = \max(a,b)$.

\subsection{Partitioning Around Medoids (PAM)}
The original PAM algorithm \cite{kaufman1987clustering, kaufman1990partitioning} first initializes the set of $k$ medoids via the BUILD step and then repeatedly performs the SWAP step to improve the loss \eqref{eqn:total-loss} until convergence.

\paragraph{BUILD:} PAM initializes a set of $k$ medoids by greedily assigning medoids one-by-one so as to minimize the overall loss \eqref{eqn:total-loss}. 
The first point added in this manner is the medoid of all $n$ points.
Given the current set of $l$ medoids $\mathcal{M}_{l} = \{m_1, \cdots, m_{l}\}$, the next point to add $m^*$ is
\begin{equation}
\label{eqn:build-next-medoid}
    \text{BUILD:~~~~}m^* = \argmin_{x \in \mathcal{X} \setminus \mathcal{M}_{l}} \frac{1}{n} \sum_{j=1}^n \left[d(x, x_j) \wedge \min_{m' \in \mathcal{M}_{l}} d(m', x_j)\right] 
\end{equation}
\paragraph{SWAP:} PAM then swaps the medoid-nonmedoid pair that would reduce the loss \eqref{eqn:total-loss} the most among all possible $k(n-k)$  such pairs.
Let $\mathcal{M}$ be the current set of $k$ medoids. Then the best medoid-nonmedoid pair $(m^*, x^*)$ to swap is
\begin{equation}
\label{eqn:next-swap}
    \text{SWAP:~~~~}(m^*, x^*) = \argmin_{(m,x) \in \mathcal{M} \times (\mathcal{X} \setminus \mathcal{M}) } \frac{1}{n} \sum_{j=1}^n \left[d(x, x_j) \wedge \min_{m' \in \mathcal{M}\setminus \{m\}} d(m', x_j) \right] 
\end{equation}
The second terms in \eqref{eqn:build-next-medoid} and \eqref{eqn:next-swap}, namely $\min_{m' \in \mathcal{M}_{l}} d(m', x_j)$ and $\min_{m' \in \mathcal{M}\setminus \{m\}} d(m', x_j)$, can be determined by caching the smallest and the second smallest distances from each point to the previous set of medoids, namely $\mathcal{M}_{l}$ in \eqref{eqn:build-next-medoid} and $\mathcal{M}$ in \eqref{eqn:next-swap}.
Therefore, in \eqref{eqn:build-next-medoid} and \eqref{eqn:next-swap}, we only need to compute the distance once for each summand.
As a result, PAM needs $O(kn^2)$ distance computations for the $k$ greedy searches in the BUILD step and $O(kn^2)$ distance computations for each SWAP iteration.



\section{\algnamenospace}
\label{sec:algo}

At the core of the PAM algorithm is the $O(n^2)$ BUILD search \eqref{eqn:build-next-medoid}, which is repeated $k$ times for initialization, and the $O(kn^2)$ SWAP search \eqref{eqn:next-swap}, which is repeated until convergence. 
We first show that both searches share a similar mathematical structure, and then show that such a structure can be optimized efficiently using a bandit-based randomized algorithm, thus giving rise to \algnamenospace. 
Rewriting the BUILD search \eqref{eqn:build-next-medoid} and the SWAP search \eqref{eqn:next-swap} in terms of the change in total loss yields
\begin{align}
    \text{BUILD:~~~~}& \argmin_{x \in \mathcal{X} \setminus \mathcal{M}_{l}} \frac{1}{n} \sum_{j=1}^n \left[ \left(d(x, x_j) - \min_{m' \in \mathcal{M}_{l}} d(m', x_j) \right)  \wedge 0\right] \label{eqn:build_search}\\
    \text{SWAP:~~~~}& \argmin_{(m,x) \in \mathcal{M} \times (\mathcal{X} \setminus \mathcal{M}) } \frac{1}{n} \sum_{j=1}^n \left[ \left(d(x, x_j) - \min_{m' \in \mathcal{M}\setminus \{m\}} d(m', x_j) \right)\wedge 0\right]  \label{eqn:swap_search}
\end{align}
One may notice that the above two problems share the following similarities.
First, both are searching over a finite set of parameters: $n-l$ points in the BUILD search and $k(n-k)$ swaps in the SWAP search. 
Second, both objective functions have the form of an average of an $O(1)$ function evaluated over a finite set of reference points. 
We formally describe the shared structure:
\begin{align} \label{eqn:genetic_optimiation}
    \text{Shared Problem:~~~~} \argmin_{x \in \mathcal{S}_{\text{tar}} } \frac{1}{\vert \mathcal{S}_{\text{ref}} \vert } \sum_{x_j \in \mathcal{S}_{\text{ref}}} g_x(x_j) 
\end{align}
for target points $\mathcal{S}_{\text{tar}} $, reference points $\mathcal{S}_{\text{ref}}$, and an objective function $g_x(\cdot)$ that depends on the target point $x$. Then both BUILD and SWAP searches can be written as instances of Problem \eqref{eqn:genetic_optimiation} with:
\begin{align}
    & \text{BUILD:~~} 
    \mathcal{S}_{\text{tar}}=\mathcal{X} \setminus \mathcal{M}_{l},~
    \mathcal{S}_{\text{ref}} = \mathcal{X},~
    g_x(x_j) = \left(d(x, x_j) - \min_{m' \in \mathcal{M}_{l}} d(m', x_j) \right)  \wedge 0, \label{eqn:build_instance}\\
    & \text{SWAP:~~} 
    \mathcal{S}_{\text{tar}}=\mathcal{M} \times (\mathcal{X} \setminus \mathcal{M}),~
    \mathcal{S}_{\text{ref}} = \mathcal{X},~
    g_x(x_j) =  \left(d(x, x_j) - \min_{m' \in \mathcal{M} \setminus \{m\}} d(m', x_j) \right)\wedge 0. \label{eqn:swap_instance}
\end{align}
Crucially, in the SWAP search, each \textit{pair} of medoid-and-non-medoid points $(m,x)$ is treated as one target point in $\mathcal{S}_{\text{tar}}$ in this new formulation.

\subsection{Adaptive search for the shared problem}
Recall that the computation of $g(x_j)$ is $O(1)$.
A naive, explicit method would require $O(\vert \mathcal{S}_{\text{tar}} \vert \vert \mathcal{S}_{\text{ref}} \vert)$ computations of $g(x_j)$ to solve Problem \eqref{eqn:genetic_optimiation}. 
However, as shown in previous works \cite{bagaria2018medoids,bagaria2018adaptive}, a randomized search would return the correct result with high confidence in $O( \vert \mathcal{S}_{\text{tar}}\vert \log  \vert \mathcal{S}_{\text{ref}} \vert)$ computations of $g(x_j)$.
Specifically, for each target $x$ in Problem \eqref{eqn:genetic_optimiation}, let $\mu_x = \frac{1}{\vert \mathcal{S}_{\text{ref}} \vert } \sum_{x_j \in \mathcal{S}_{\text{ref}}} g_x(x_j)$ denote its objective function. Computing $\mu_x$ exactly takes $O(\vert \mathcal{S}_{\text{ref}} \vert)$ computations of $g(x_j)$, but we can instead estimate $\mu_x$ with fewer computations by drawing $J_1,J_2,...,J_{n'}$ independent samples uniformly with replacement from $[|\mathcal{S_{\text{ref}}}|]$.
Then, $E[g(x_{J_i})] = \mu_x$ and $\mu_x$ can be estimated as $\hat{\mu}_x = \frac{1}{n'} \sum_{i=1}^{n'} g(x_{J_i})$, where $n'$ determines the estimation accuracy. 
To estimate the solution to Problem \eqref{eqn:genetic_optimiation} with high confidence, we can then choose to sample different targets in $\mathcal{S}_{\text{tar}}$ to different degrees of accuracy. 
Intuitively, promising targets with small values of $\mu_x$ should be estimated with high accuracy, while less promising ones can be discarded without being evaluated on too many reference points. 


The specific adaptive estimation procedure is described in Algorithm \ref{alg:bandit_based_search}. 
It can be viewed as a batched version of the conventional UCB algorithm \cite{lai1985asymptotically,zhang2019adaptive} combined with successive elimination \cite{successiveelimination}, and is straightforward to implement.
Algorithm \ref{alg:bandit_based_search} uses the set $\mathcal{S}_{\text{solution}}$ to track all potential solutions to Problem \eqref{eqn:genetic_optimiation}; $\mathcal{S}_{\text{solution}}$ is initialized as the set of all target points $\mathcal{S}_{\text{tar}}$. We will assume that, for a fixed target point $x$ and a randomly sampled reference point $x_J$, the random variable $Y = g_x(x_J)$ is $\sigma_x$-sub-Gaussian for some known parameter $\sigma_x$. Then, for each potential solution $x\in \mathcal{S}_{\text{solution}}$,  Algorithm \ref{alg:bandit_based_search} maintains its mean objective estimate $\hat{\mu}_x$ and confidence interval $C_x$, where $C_x$ depends on the exclusion probability $\delta$ as well as the parameter $\sigma_x$. We discuss the sub-Gaussianity parameters and possible relaxations of this assumption in Sections \ref{sec:theory} and \ref{sec:discussion} and Appendix \ref{app:relaxation}.

In each iteration, a new batch of reference points $\mathcal{S}_{\text{ref\_batch}}$ is evaluated for all potential solutions in $\mathcal{S}_{\text{solution}}$, making the estimate $\hat{\mu}_x$ more accurate. 
Based on the current estimate, if a target's lower confidence bound $\hat{\mu}_x - C_x$ is greater than the upper confidence bound of the most promising target $\min_{y}(\hat{\mu}_{y} + C_{y})$, we remove it from 
$\mathcal{S}_{\text{solution}}$. This process continues until there is only one point in $\mathcal{S}_{\text{solution}}$ or until we have sampled more reference points than in the whole reference set. In the latter case, we know that the difference between the remaining targets in $\mathcal{S}_{\text{solution}}$ is so subtle that an exact computation is more efficient. We then compute those targets' objectives exactly and return the best target in the set. 


\begin{algorithm}[t]

\caption{
\texttt{Adaptive-Search} (
$\mathcal{S}_{\text{tar}},
\mathcal{S}_{\text{ref}},
g_x(\cdot)$,
$B$,
$\delta$,
$\sigma_x$
) \label{alg:bandit_based_search}}
\begin{algorithmic}[1]
\State $\mathcal{S}_{\text{solution}} \leftarrow \mathcal{S}_{\text{tar}}$ \Comment{Set of potential solutions to Problem \eqref{eqn:genetic_optimiation}}
\State $n_{\text{used\_ref}} \gets 0$  \Comment{Number of reference points evaluated}
\State For all $x \in  \mathcal{S}_{\text{tar}}$, set $\hat{\mu}_x \leftarrow 0$, $C_x \leftarrow \infty$  \Comment{Initial mean and confidence interval for each arm}

\While{$n_{\text{used\_ref}} < \vert \mathcal{S}_{\text{ref}} \vert $ and $|\mathcal{S}_{\text{solution}}| > 1$} 
        \State Draw a batch samples of size $B$ with replacement from reference $\mathcal{S}_{\text{ref\_batch}} \subset \mathcal{S}_{\text{ref}}$ 
        \ForAll{$x \in \mathcal{S}_{\text{solution}} $}
            \State $\hat{\mu}_x \leftarrow \frac{ n_{\text{used\_ref}} \hat{\mu}_x + \sum_{y \in \mathcal{S}_{\text{ref\_batch}}} g_x(y)}{n_{\text{used\_ref}} + B }$ \Comment{Update running mean}
            \State $C_x \gets \sigma_x \sqrt{  \frac{ \log(\frac{1}{\delta}) } {n_{\text{used\_ref}} + B }}$
            \Comment{Update confidence interval}        
        \EndFor
    \State $\mathcal{S}_{\text{solution}} \leftarrow \{x : \hat{\mu}_x - C_x \leq \min_{y}(\hat{\mu}_{y} + C_{y})\}$ \Comment{Remove points that can no longer be solution}
    \State $n_{\text{used\_ref}} \leftarrow n_{\text{used\_ref}} + B$
\EndWhile
\If{$\vert \mathcal{S}_{\text{solution}} \vert$ = 1}
    \State \textbf{return} $x^* \in \mathcal{S}_{\text{solution}}$
\Else
    \State Compute $\mu_x$ exactly for all $x \in \mathcal{S}_{\text{solution}}$
    \State \textbf{return} $x^* = \argmin_{x \in \mathcal{S}_{\text{solution}}} \mu_x$
\EndIf
\end{algorithmic}
\end{algorithm}

\subsection{Algorithmic details \label{subsec:algdetails}}
\label{A0}

\textbf{Estimation of each $\sigma_x$:} \algname uses Algorithm \ref{alg:bandit_based_search} in both the BUILD step and each SWAP iteration, with input parameters specified in \eqref{eqn:build_instance} and \eqref{eqn:swap_instance}. In practice, $\sigma_x$ is not known \emph{a priori} and we estimate $\sigma_x$ for each $x \in |\mathcal{S}_{\text{tar}}|$ from the data. In the first batch of sampled reference points in Algorithm \ref{alg:bandit_based_search}, we estimate each $\sigma_x$ as:
\begin{equation}
\label{eqn:sigma_est}
    \sigma_x = {\rm STD}_{y \in \mathcal{S}_{\text{ref\_batch}}}g_x(y)
\end{equation}
where ${\rm STD}$ denotes standard deviation. Intuitively, this allows for smaller confidence intervals in later iterations, especially in the BUILD step, when we expect the average arm returns to become smaller as we add more medoids (since we are taking the minimum over a larger set on the RHS of Eq.~\eqref{eqn:build-next-medoid}). We also allow for arm-dependent $\sigma_x$, as opposed to a fixed global $\sigma$, which allows for narrower confidence intervals for arms whose returns are heavily concentrated (e.g., distant outliers). 
Empirically, this results in significant speedups and results in fewer arms being computed exactly (Line 14 in Algorithm \ref{alg:bandit_based_search}). In all experiments, the batch size $B$ is set to 100 and the error probability $\delta$ is set to $\delta = \frac{1}{1000\vert \mathcal{S}_{\text{tar}} \vert}$. Empirically, these values of batch size and this setting of $\delta$ are such that \algname recovers the same results in PAM in almost all cases.

\textbf{Combination with FastPAM1:} We also combine \algname with the FastPAM1 optimization \cite{schubert2019faster}. We discuss this optimization in Appendix \ref{A1}.

\section{Analysis of the Algorithm \label{sec:theory}}

The goal of \algname
is to track the optimization trajectory of the standard PAM algorithm, ultimately identifying the same set of $k$ medoids with high probability.
In this section, we formalize this statement
and provide bounds on the number of distance computations required by \algnamenospace. We begin by considering a single call to Algorithm \ref{alg:bandit_based_search} and showing it returns the correct result with high probability. We then repeatedly apply Algorithm \ref{alg:bandit_based_search} to track PAM's optimization trajectory throughout the BUILD and SWAP steps.


Consider a single call to Algorithm \ref{alg:bandit_based_search} and suppose $x^* = \argmin_{x \in \Star} \mu_x$ is the optimal target point.
For another target point $x \in \Star$,
let $\Delta_x \vcentcolon = \mu_x - \mu_{x^*}$.
To state the following results, we will assume that, for a fixed target point $x$ and a randomly sampled reference point $x_J$,
the random variable $Y = g_x(x_J)$ is $\sigma_x$-sub-Gaussian
for some known parameter $\sigma_x$. 
In practice, one can estimate each $\sigma_x$ by performing a small number of distance computations as described in Section \ref{subsec:algdetails}.
Allowing $\sigma_x$ to be estimated separately for each arm is beneficial in practice, as discussed in Section \ref{sec:discussion}.
With these assumptions, the following theorem is proved in Appendix \ref{app:thmproof}:




\begin{theorem} \label{thm:specific}
For $\delta = n^{-3}$, with probability at least $1-\tfrac{2}{n}$, Algorithm \ref{alg:bandit_based_search}
returns the correct solution to \eqref{eqn:build_search} (for a BUILD step) or \eqref{eqn:swap_search} (for a SWAP step),
using a total of $M$ distance computations, where
\aln{
E[M] \leq 4n + \sum_{x \in \X}  \min \left[ \frac{12}{\Delta_x^2} \left(\sigma_x+\sigma_{x^*} \right)^2 \log n + B, 2n \right].
}
\end{theorem}





Intuitively, Theorem \ref{thm:specific} states that with high probability, each step of \algname returns the same result as PAM.
For the general result, we assume that the data is generated in a way such that the mean rewards $\mu_x$ follow a sub-Gaussian distribution (see Section \ref{sec:discussion} for a discussion). 
Additionally, we assume that both PAM and \algname place a hard constraint $T$ on the maximum number of SWAP iterations that are allowed. 
Informally, as long as \algname finds the correct solution to the search problem \eqref{eqn:build_search} at each BUILD step and to the search problem \eqref{eqn:swap_search} at each SWAP step, it will reproduce the sequence of BUILD and SWAP steps of PAM identically and return the same set of final medoids. We formalize this statement with Theorem \ref{thm:nlogn}, and discuss the proof in Appendix \ref{app:thmproof}.
When the number of desired medoids $k$ is a constant 
and the number of allowed SWAP steps is small (which is often sufficient in practice, as discussed in Section \ref{sec:discussion}), Theorem \ref{thm:nlogn} implies that only $O(n \log n)$ distance computations are necessary to reproduce the results of PAM with high probability.

%



\begin{theorem} \label{thm:nlogn}
If \algname is run on a dataset $\X$ with $\delta = n^{-3}$, then it returns the same set of $k$ medoids as PAM with probability $1-o(1)$. 
Furthermore, 
the total number of distance computations $M_{\rm total}$ required satisfies
\aln{
E[M_{\rm total}] = O\left( n \log n \right).
}
\end{theorem}




\textbf{Remark 1:} While the limit on the maximum number of swap steps, $T$, may seem restrictive, it is not uncommon to place a maximum number of iterations on iterative algorithms. Furthermore, $T$ has been observed  empirically to be $O(k)$ \cite{schubert2019faster}, consistent with our experiments in Section \ref{sec:exps}.

\textbf{Remark 2:} We note that $\delta$ is a hyperparameter governing the error rate. It is possible to prove results analogous to Theorems \ref{thm:specific} and \ref{thm:nlogn} for arbitrary $\delta$; we discuss this in Appendix \ref{app:thmproof}.

\textbf{Remark 3:} Throughout this work, we have assumed that evaluating the distance between two points is $O(1)$ rather than $O(d)$, where $d$ is the dimensionality of the datapoints. If we were to include this dependence explicitly, we would have $E[M_{\rm total}] = O(d n \log n)$ in Theorem \ref{thm:nlogn}. 
We discuss improving the scaling with $d$ in Appendix \ref{appendix:scalingwithd}, and the explicit dependence on $k$ in Appendix \ref{appendix:scalingwithk}.






\section{Empirical Results \label{sec:exps}}

\textbf{Setup:} As discussed in Section \ref{sec:intro}, PAM has been empirically observed to produce the best results for the $k$-medoids problem in terms of clustering quality. Other existing algorithms can generally be divided into several classes: those that agree exactly with PAM (e.g. FastPAM1), those that do not agree exactly with PAM but provide comparable results (e.g. FastPAM) and other randomized algorithms that sacrifice clustering quality for runtime.
In Subsection \ref{subsec:loss}, we show that \algname returns the same results as PAM, thus matching the state-of-the-art in clustering quality, and also results in better or comparable final loss when compared to other popular $k$-medoids clustering algorithms, including FastPAM \cite{schubert2019faster}, CLARANS \cite{ng2002clarans}, and Voronoi Iteration \cite{park2009simple}.
In Subsection \ref{subsec:scaling}, we demonstrate that \algname scales almost linearly in the number of samples $n$ for all datasets and all metrics considered, which is superior to the quadratic scaling of PAM, FastPAM1, and FastPAM. Combining these observations, we conclude that \algname matches state-of-the-art algorithms in clustering quality, while reaching its solutions much faster.
In the experiments, each parameter setting was repeated $10$ times with data subsampled from the original dataset and 95\% confidence intervals are provided. 

\paragraph{Datasets:} We run experiments on three real-world datasets to validate the behavior of \algnamenospace, %
all of which are publicly available. The MNIST dataset \cite{lecun1998gradient} consists of 70,000 black-and-white images of handwritten digits, where each digit is represented as a 784-dimensional vector. On MNIST, We consider two distance metrics: $l_2$ distance and cosine distance.
The scRNA-seq dataset contains the gene expression levels of 10,170 different genes in each of 40,000 cells after standard filtering. On scRNA-seq, we consider $l_1$ distance, which is recommended \cite{ntranos2016fast}.
The HOC4 dataset from Code.org \cite{hoc4dataset} consists of 3,360 unique solutions to a block-based programming exercise. Solutions to the programming exercise are represented as abstract syntax trees (ASTs), and we consider the tree edit distance \cite{zss} to quantify similarity between solutions. 

\begin{center}
\begin{figure}[ht]
    \begin{subfigure}{.5\textwidth}
        \centering
        \includegraphics[width=\linewidth]{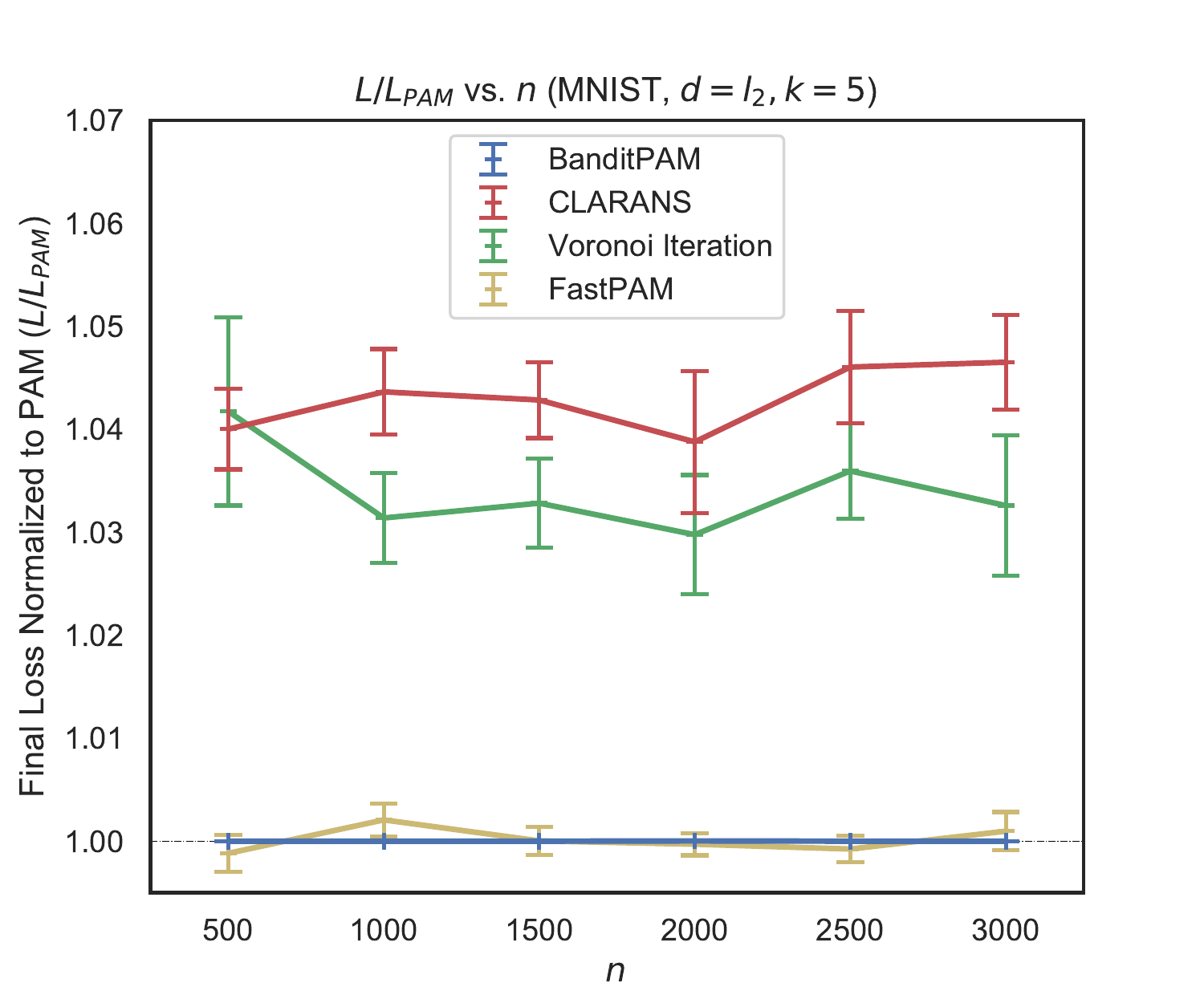} 
        \caption{}
    \end{subfigure}
    \begin{subfigure}{.5\textwidth}
      \centering
      \includegraphics[width=\linewidth]{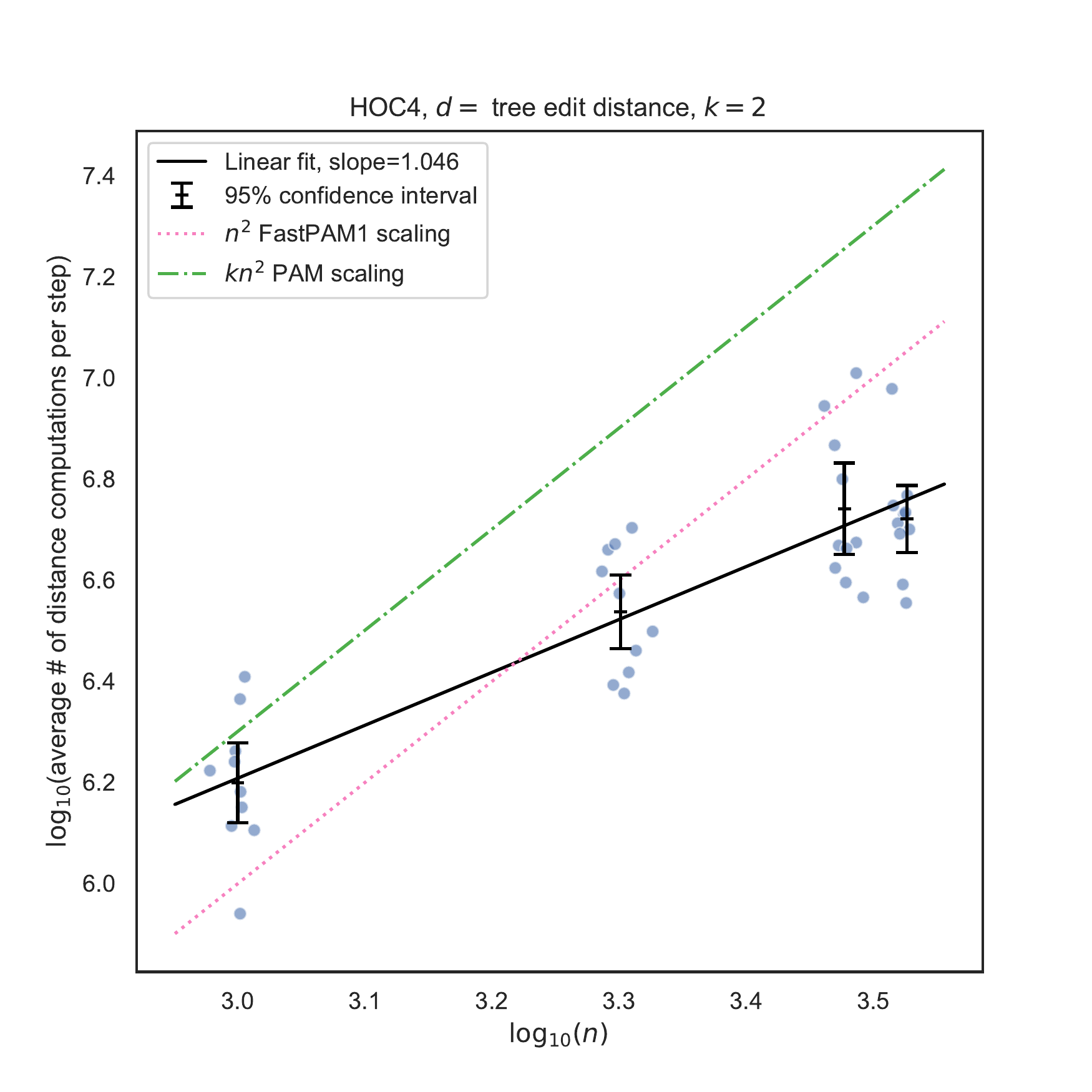}   
      \caption{}
    \end{subfigure}
    \caption{(a) Clustering loss relative to PAM loss. Data is subsampled from MNIST, sample size $n$ varies from $500$ to $3000$, $k = 5$, and 95\% confidence intervals are provided. \algname always returns the same solution as PAM and hence has loss ratio exactly $1$, as does FastPAM1 (omitted for clarity). FastPAM also demonstrates comparable final loss, while the other two algorithms are significantly worse. (b) Average number of distance evaluations per iteration vs. sample size $n$ for HOC4 and tree edit distance, with $k = 2$, on a log-log plot. Reference lines for PAM and FastPAM1 are also shown. \algname scales better than PAM and FastPAM1 and is significantly faster for large datasets.}
    \label{fig:losses}
\end{figure}
\end{center}

\begin{figure}[ht]
    \begin{subfigure}{.5\textwidth}
      \centering
      \includegraphics[width=\linewidth]{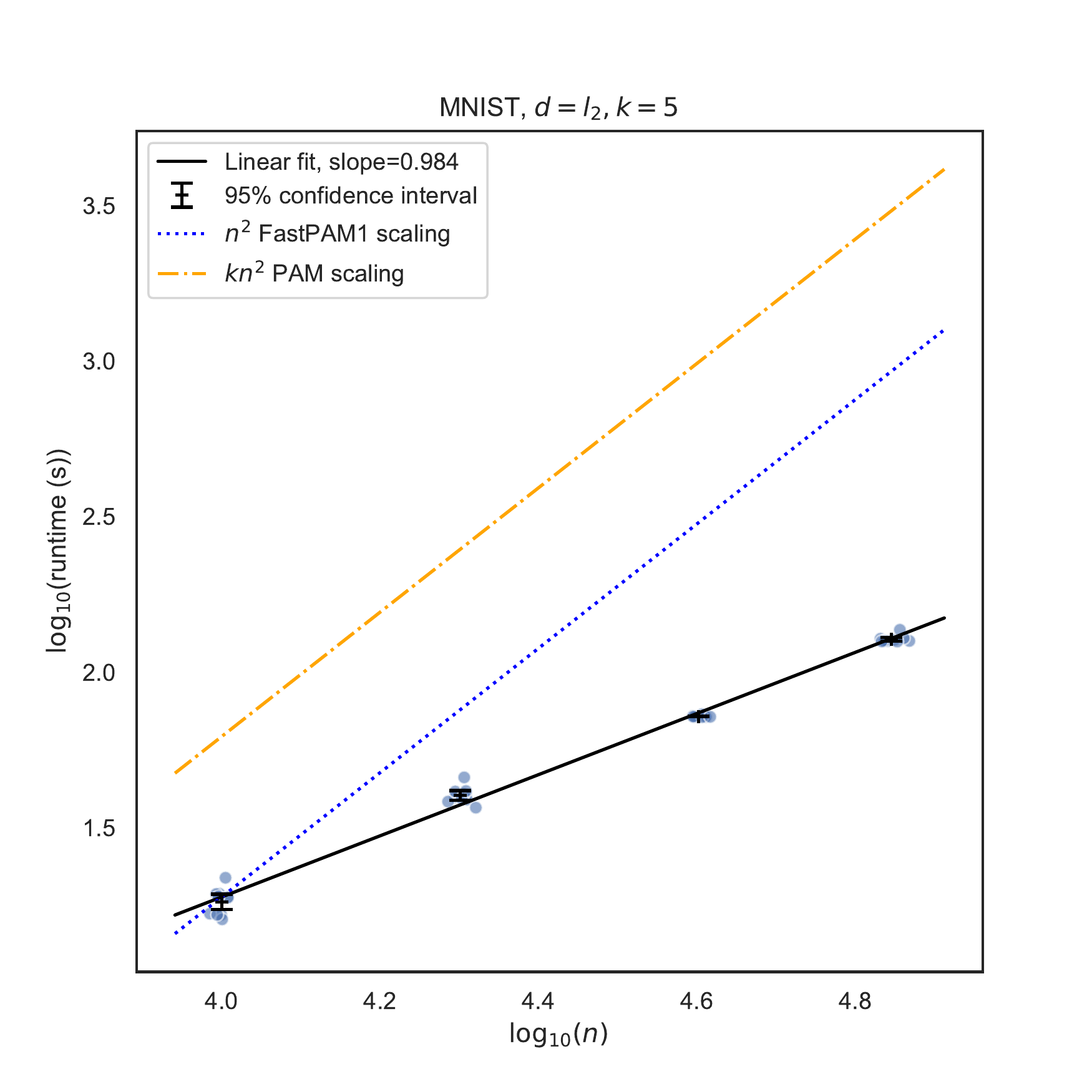}  
      \caption{}
    \end{subfigure}
    \begin{subfigure}{.5\textwidth}
      \centering
      \includegraphics[width=\linewidth]{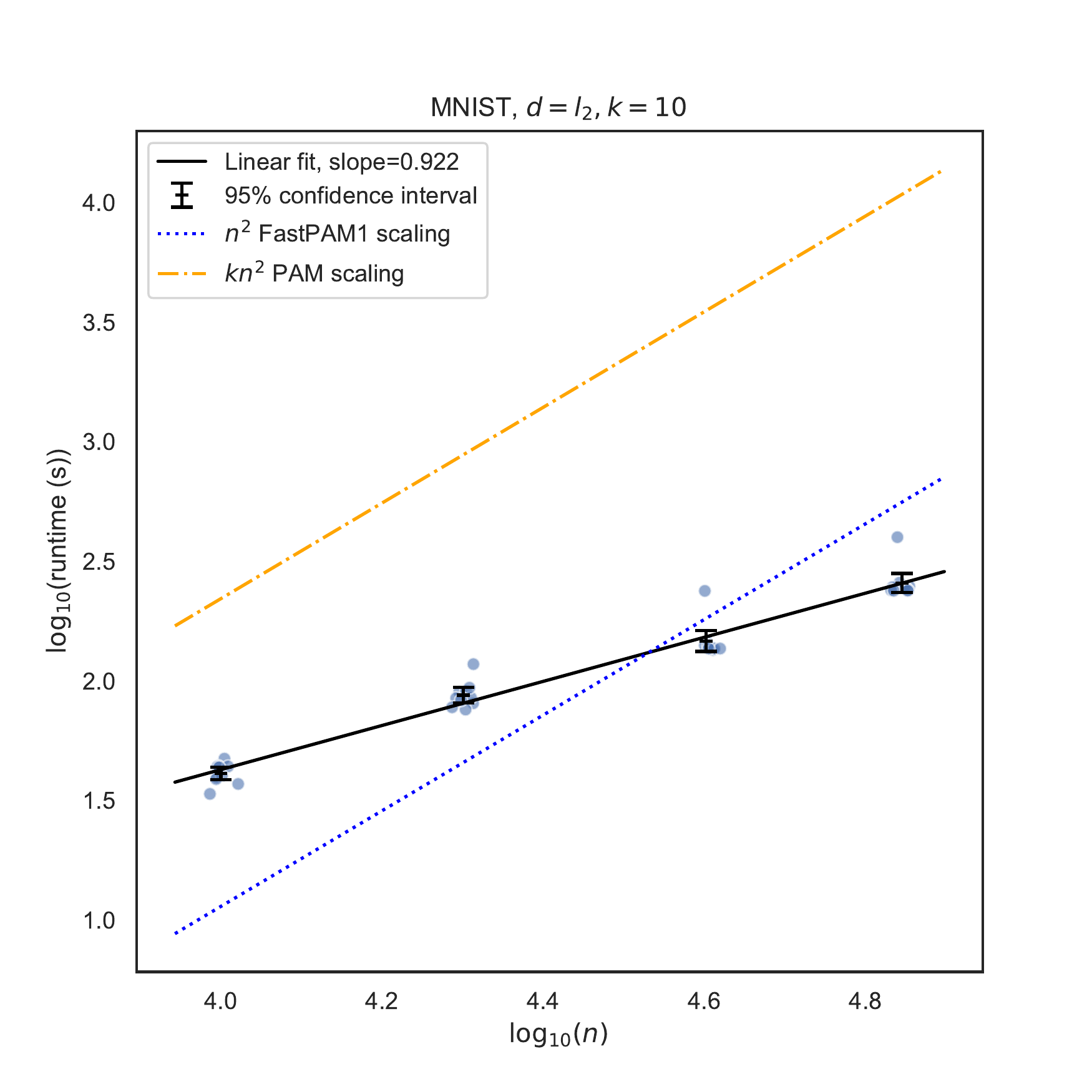}   
      \caption{}
    \end{subfigure}
    \caption{Average runtime per iteration vs. sample size $n$ for MNIST and $l_2$ distance with (a) $k=5$ and (b) $k=10$, on a log-log scale. Lines of best fit (black) are plotted, as are reference lines demonstrating the expected scaling of PAM and FastPAM1.}
    \label{fig:mnist-l2}
\end{figure}

\subsection{Clustering/loss quality \label{subsec:loss}}


Figure \ref{fig:losses} (a) shows the relative losses of algorithms with respect to the loss of PAM. 
\algname and three other baselines, namely FastPAM \cite{schubert2019faster}, CLARANS \cite{ng2002clarans}, and Voronoi Iteration \cite{park2009simple}. 
We clarify the distinction between FastPAM and FastPAM1; both are $O(n^2)$ in each SWAP step but FastPAM1 is guaranteed to return the same solution as PAM while FastPAM is not. 
In these experiments, \algname returns the same solution as PAM and hence has loss ratio $1$. FastPAM has a comparable performance, while the other two algorithms are significantly worse.





\subsection{Scaling with \texorpdfstring{$n$}{Lg} for different datasets, distance metric, and \texorpdfstring{$k$}{Lg} values \label{subsec:scaling}}

\begin{figure}[ht]
    \begin{subfigure}{.5\textwidth}
      \centering
      \includegraphics[width=\linewidth]{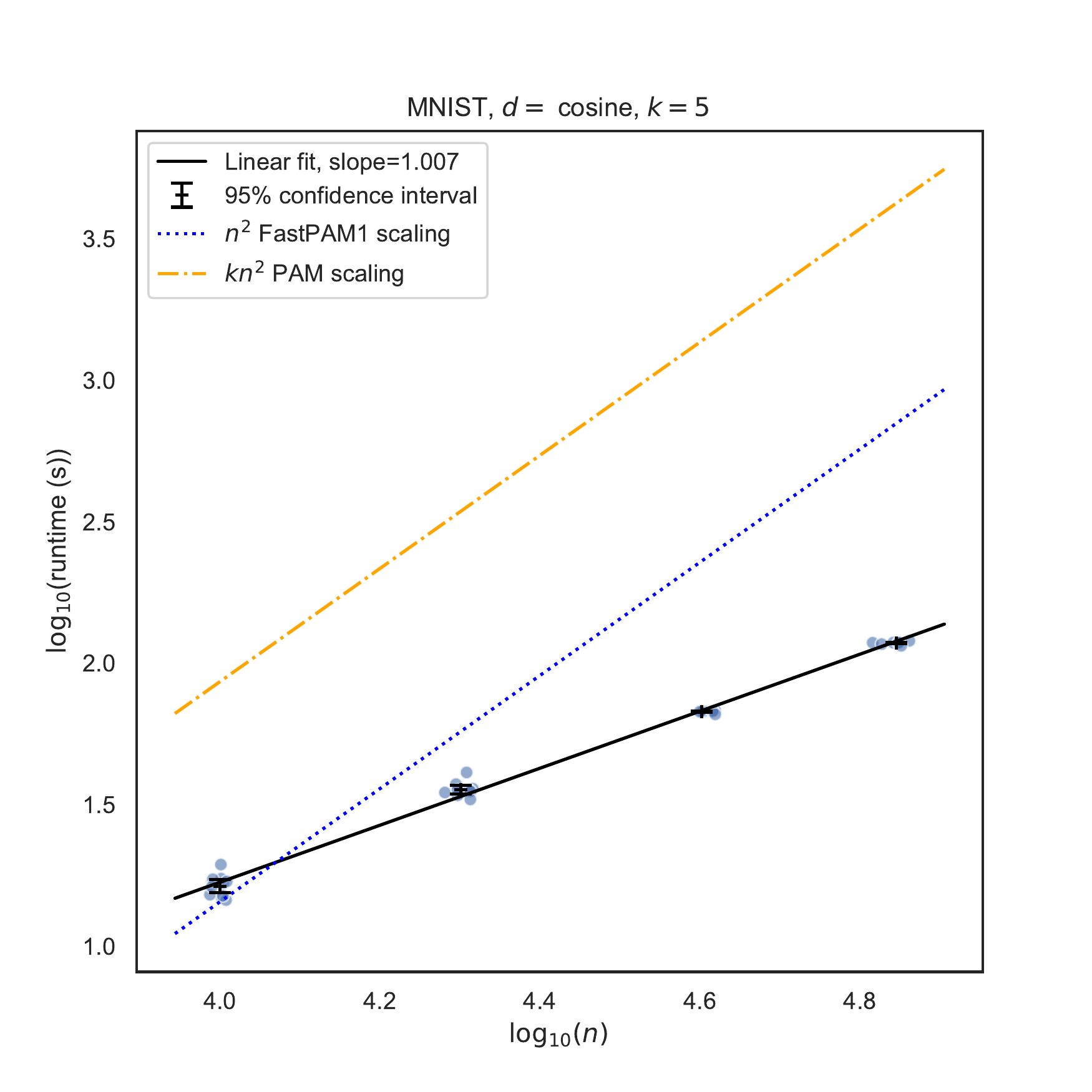}  
      \caption{}
    \end{subfigure}
    \begin{subfigure}{.5\textwidth}
      \centering
      \includegraphics[width=\linewidth]{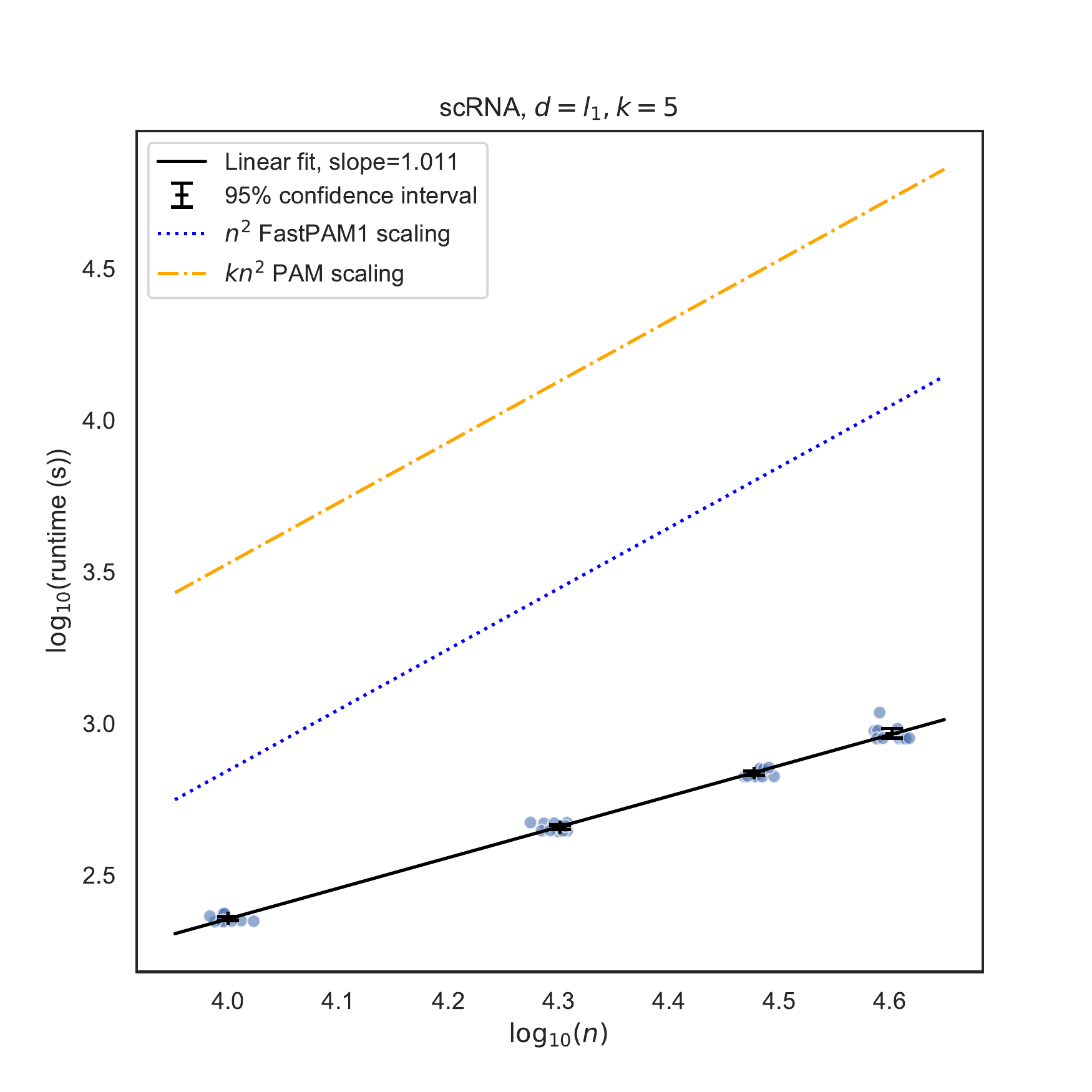}   
      \caption{}
    \end{subfigure}
    \caption{Average runtime per iteration vs. sample size $n$, for (a) MNIST and cosine distance and (b) scRNA-seq and $l_1$ distance, with $k = 5$. Lines of best fit (black) are plotted, as are reference lines demonstrating the expected scaling of PAM and FastPAM1.}
    \label{fig:mnist-scrna}
\end{figure}


We next consider the runtime per iteration of \algnamenospace, especially in comparison to PAM and FastPAM1. To calculate the runtime per iteration of \algnamenospace, we divide the total wall clock time by the number of SWAP iterations plus 1, where each SWAP step has expected complexity $O(kn\log n)$ and the plus 1 accounts for the $O(kn\log n)$ complexity of all $k$ BUILD steps. 
Figure \ref{fig:mnist-l2} demonstrates the runtime per iteration of \algname versus $n$ on a log-log plot. The slopes for the lines of best fit for (a) $k=5$ and (b) $k=10$ are 0.984 and 0.922, respectively, indicating the scaling is linear in $n$ for different values of $k$.

Figure \ref{fig:mnist-scrna} demonstrates the runtime per iteration of \algname for other datasets and metrics. On MNIST with cosine distance (a), the slope of the line of best fit is 1.007. On the scRNA-seq dataset with $l_1$ distance (b), the slope of the line of best fit is 1.011. These results validate our theory that \algname takes almost linear number of distance evaluations per iteration for different datasets and different distance metrics.

Because the exact runtime of \algname and other baselines depends on implementation details such as programming language, we also analyze the number of distance evaluations required by each algorithm. Indeed, a profile of \algname program reveals that it spends over 98\% of its wall clock time computing distances; as such, the number of distance evaluations provides a reasonable proxy for complexity of \algnamenospace. For the other baselines PAM and FastPAM1, the number of distance evaluations is expected to be exactly $kn^2$ and $n^2$, respectively, in each iteration. Figure \ref{fig:losses} (b) demonstrates the number of distance evaluations per iteration of \algname with respect to $n$. The slope of the line of best fit on the log-log plot is 1.046, which again indicates that \algname scales linearly in dataset size even for more exotic objects and metrics, such as trees and tree edit distance.







\section{Discussion and Conclusions \label{sec:discussion}}

In this work, we proposed \algnamenospace, a randomized algorithm for the $k$-medoids problem that matches state-of-the-art approaches in clustering quality while achieving a reduction in complexity from $O(n^2)$ to $O(n\log n)$ under certain assumptions. In our experiments, the randomly sampled distances have an empirical distribution similar to a Gaussian (Appendix Figures \ref{fig:sigma_ex_MNIST}-\ref{fig:sigma_ex_SCRNAPCA}), justifying the sub-Gaussian assumption in Section \ref{sec:theory}.
We also observe that the the sub-Gaussian parameters are different across steps and target points (Appendix Figures \ref{fig:MNIST_sigmas_example}), justifying the adaptive estimation of the sub-Gaussianity parameters in Subsection \ref{subsec:algdetails}.
Additionally, the empirical distribution of the true arm parameters (Appendix Figure \ref{fig:mu_dist}) appears to justify the distributional assumption of $\mu_x$s in Section \ref{sec:theory}.






\section*{Broader Impact \label{sec:impact}}

\algname accelerates finding solutions to the $k$-medoids problem while producing comparable -- and usually equivalent -- final cluster assignments. Our work enables the discovery of high-quality medoid assignments in very large datasets, including some on which prior algorithms were prohibitively expensive. A potential negative consequence of this is that practitioners may be incentivized to gather and store larger amounts of data now that it can be meaningfully processed, in a phenomenon more generally described as induced demand \cite{induceddemand}. This incentive realignment could potentially result in negative externalities such as an increase in energy consumption and carbon footprints.

Our application to the HOC4 dataset suggests a method for scaling personalized feedback to individual students in online courses. If limited resources are available, instructors can choose to provide feedback on just the \textit{medoids} of submitted solutions instead of exhaustively providing feedback on \textit{every} unique solution, of which there may be several thousand. Instructors can then refer individual students to the feedback provided for their closest medoid. We anticipate that this approach can be applied generally for students of Massive Open Online Courses (MOOCs), thereby enabling more equitable access to education and personalized feedback for students.

We also anticipate, however, that \algname will enable several beneficial applications in other fields such as biomedicine and and fairness. For example, the evolutionary pathways of infectious diseases could possibly be constructed from the medoids of genetic sequences available at a given point in time, if prior temporal information about these sequences' histories is not available. Similarly, the medoids of patients infected in a disease outbreak may elucidate the origins of outbreaks, as did prior analyses of cholera outbreaks using Voronoi Iteration \cite{cholera}. As discussed in Section \ref{sec:discussion}, our application to the HOC4 data also demonstrates the utility of \algname in online education. In particular, especially with recent interest in online learning, we hope that our work will improve the quality of online learning for students worldwide.

\begin{ack}
\label{ack}
This research was funded in part by JPMorgan Chase \& Co.  Any views or opinions expressed herein are solely those of the authors listed, and may differ from the views and opinions expressed by JPMorgan Chase \& Co. or its affiliates. This material is not a product of the Research Department of J.P. Morgan Securities LLC. This material should not be construed as an individual recommendation for any particular client and is not intended as a recommendation of particular securities, financial instruments or strategies for a particular client.  This material does not constitute a solicitation or offer in any jurisdiction.  
M.Z. was also supported by NIH grant R01 MH115676.
We would like to thank Eric Frankel for help with the C++ implementation of \algnamenospace.

\end{ack}

\bibliographystyle{plain}
\bibliography{refs}


\clearpage
\section*{Appendix}
\label{appendix}

\renewcommand\thefigure{\arabic{figure}}
\setcounter{figure}{0}      
\setcounter{section}{0}

\renewcommand{\figurename}{Appendix Figure}

\section{Additional Discussions}
\subsection{FastPAM1 optimization}
\label{A2}

Algorithm \ref{alg:bandit_based_search} can also be combined with the FastPAM1 optimization from \cite{schubert2019faster} to reduce the computation in each SWAP iteration. 
For a given candidate swap $(m, x)$, we rewrite $g_{(m,x)}(x_j)$ from Equation \eqref{eqn:swap_instance} as:
\begin{equation}
    \label{eqn:fastpam1-trick}
    g_{m,x}(x_j) = - d_1(x_j) + \mathbbm{1}_{x_j \notin \mathcal{C}_m}\min[d_1(x_j),d(x,x_j)] + \mathbbm{1}_{x_j \in \mathcal{C}_m}\min[d_2(x_j), d(x, x_j)]
\end{equation}
where $\mathcal{C}_m$ denotes the set of points whose closest medoid is $m$ and $d_1(x_j)$ and $d_2(x_j)$ are the distance from $x_j$ to its nearest and second nearest medoid, respectively, before the swap is performed. 
We cache the values $d_1(x_j), d_2(x_j)$, and the cluster assignments $\mathcal{C}_m$ so that Equation \eqref{eqn:fastpam1-trick} no longer depends on $m$ and instead depend only on $\mathbbm{1}_{\{x_j \in \mathcal{C}_m \}}$, which is cached. This allows for an $O(k)$ speedup in each SWAP iteration since we do not need to recompute Equation \ref{eqn:fastpam1-trick} for each of the $k$ distinct medoids (values of $m$).

\subsection{Value of re-estimating each \texorpdfstring{$\sigma_x$}{Lg}}
\label{A1}




\begin{figure}[h!]
    \centering
    \includegraphics[scale=0.5]{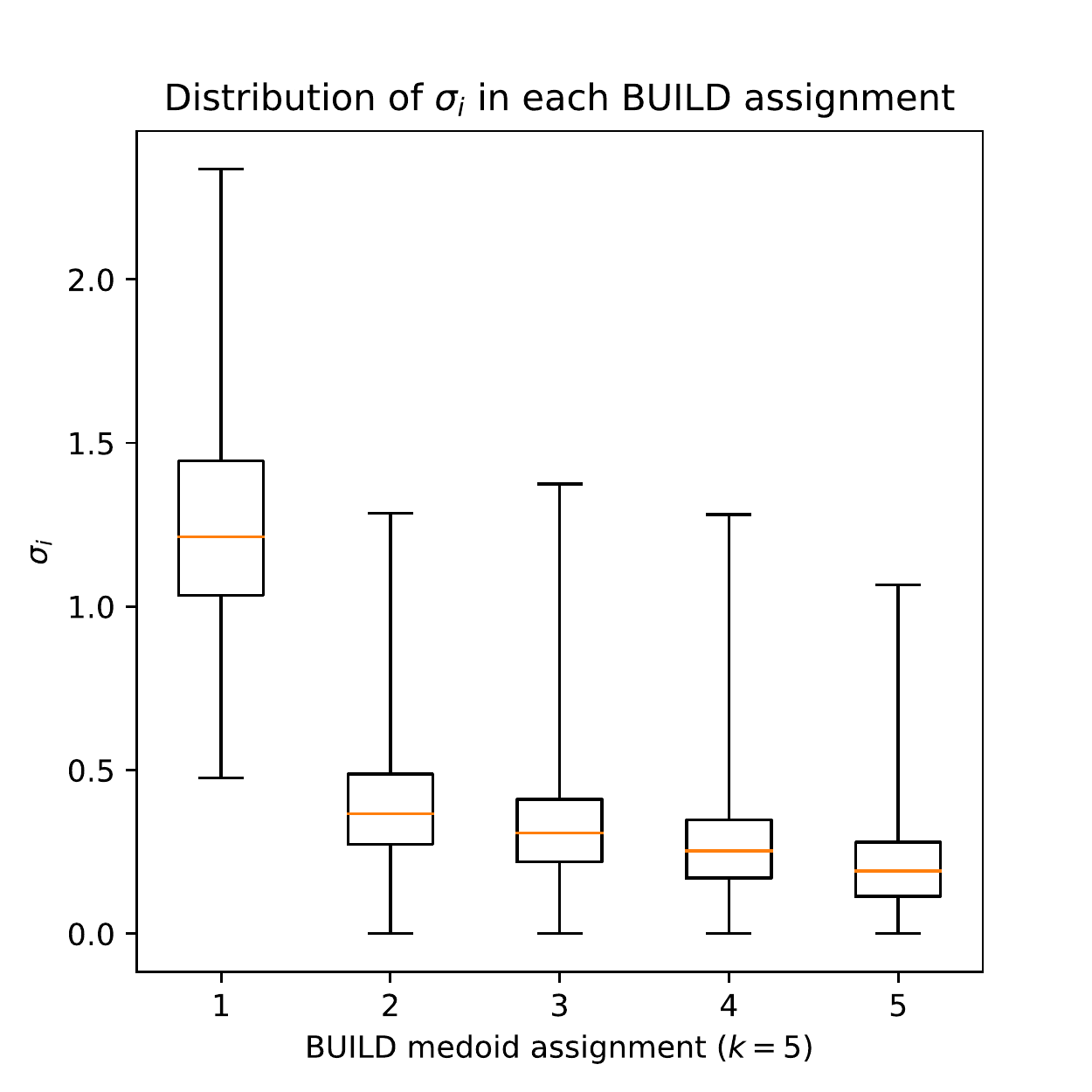}
    \caption{Boxplot showing the min, max, and each quartile for the set of all $\sigma_x$ estimates for the full MNIST dataset, in the BUILD step. } 
    \label{fig:MNIST_sigmas_example}
\end{figure}

The theoretical results in Section \ref{sec:theory} and empirical results in Section \ref{sec:exps} suggest that \algname scales almost linearly in dataset size for a variety of real-world datasets and commonly used metrics. One may also ask if Lines 7-8 of Algorithm \ref{alg:bandit_based_search}, in which we re-estimate each $\sigma_x$ from the data, are necessary.
In some sense, we treat the set of \{$\sigma_x$\} as adaptive in two different ways: $\sigma_x$ is calculated on a \textit{per-arm} basis (hence the subscript $x$), as well recalculated in each BUILD and SWAP iteration.
In practice, we observe that re-estimating each $\sigma_x$ for each sequential call to Algorithm \ref{alg:bandit_based_search} significantly improves the performance of our algorithm. Figure \ref{fig:MNIST_sigmas_example} describes the distribution of estimate $\sigma_x$ for the MNIST data at different stages of the BUILD step. The median $\sigma_x$ drops dramatically after the first medoid has been assigned and then steadily decreases, as indicated by the orange lines, and suggests that each $\sigma_x$ should be recalculated at every assignment step. Furthermore, the whiskers demonstrate significant variation amongst the $\sigma_x$ in a given assignment step and suggest that having arm-dependent $\sigma_x$ parameters is necessary. Without these modifications to our algorithm, we find that the confidence intervals used by \algname (Line 8) are unnecessarily large and cause computation to be expended needlessly as it becomes harder to identify the best target points. Intuitively, this is due to the much larger confidence intervals that make it harder to distinguish between arms' mean returns.
For a more detailed discussion of the distribution of $\sigma_x$ and examples where the assumptions of Theorem \ref{thm:nlogn} are violated, we refer the reader to Appendix \ref{A3}.

\subsection{Violation of distributional assumptions}
\label{A3}

In this section, we investigate the robustness of \algname to violations of the assumptions in Theorem \ref{thm:nlogn} on an example dataset and provide intuitive insights into the degradation of scaling. We create a new dataset from the scRNA dataset by projecting each point onto the top 10 principal components of the dataset; we call the dataset of projected points scRNA-PCA. Such a transformation is commonly used in prior work; the most commonly used distance metric between points is then the $l_2$ distance \cite{scrnabestpractices}.

\begin{figure}[ht]
\begin{subfigure}{.5\textwidth}
  \centering
  \includegraphics[width=\linewidth]{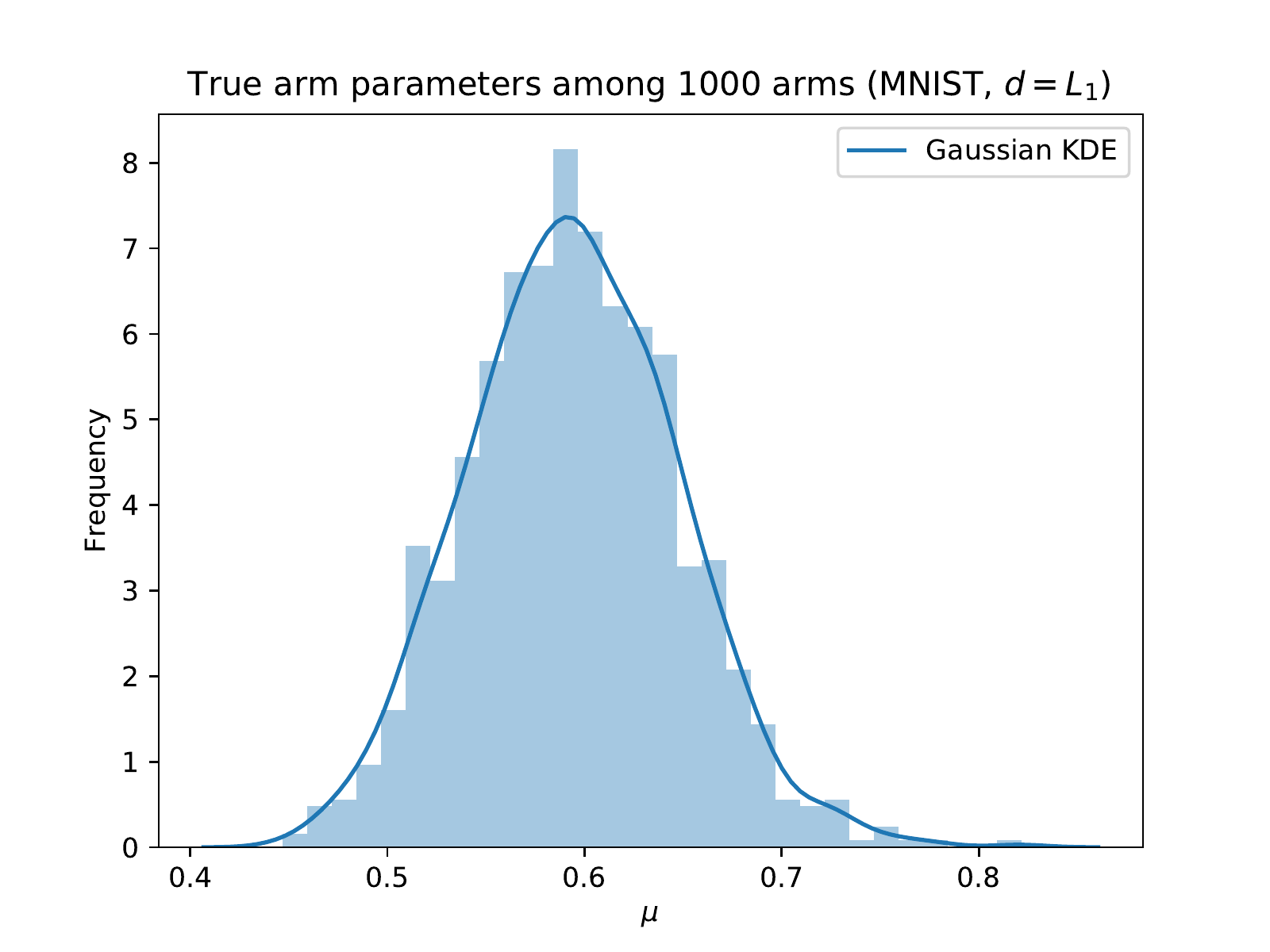}  
  \label{fig:mu_dist1}
\end{subfigure}
\begin{subfigure}{.5\textwidth}
  \centering
  \includegraphics[width=\linewidth]{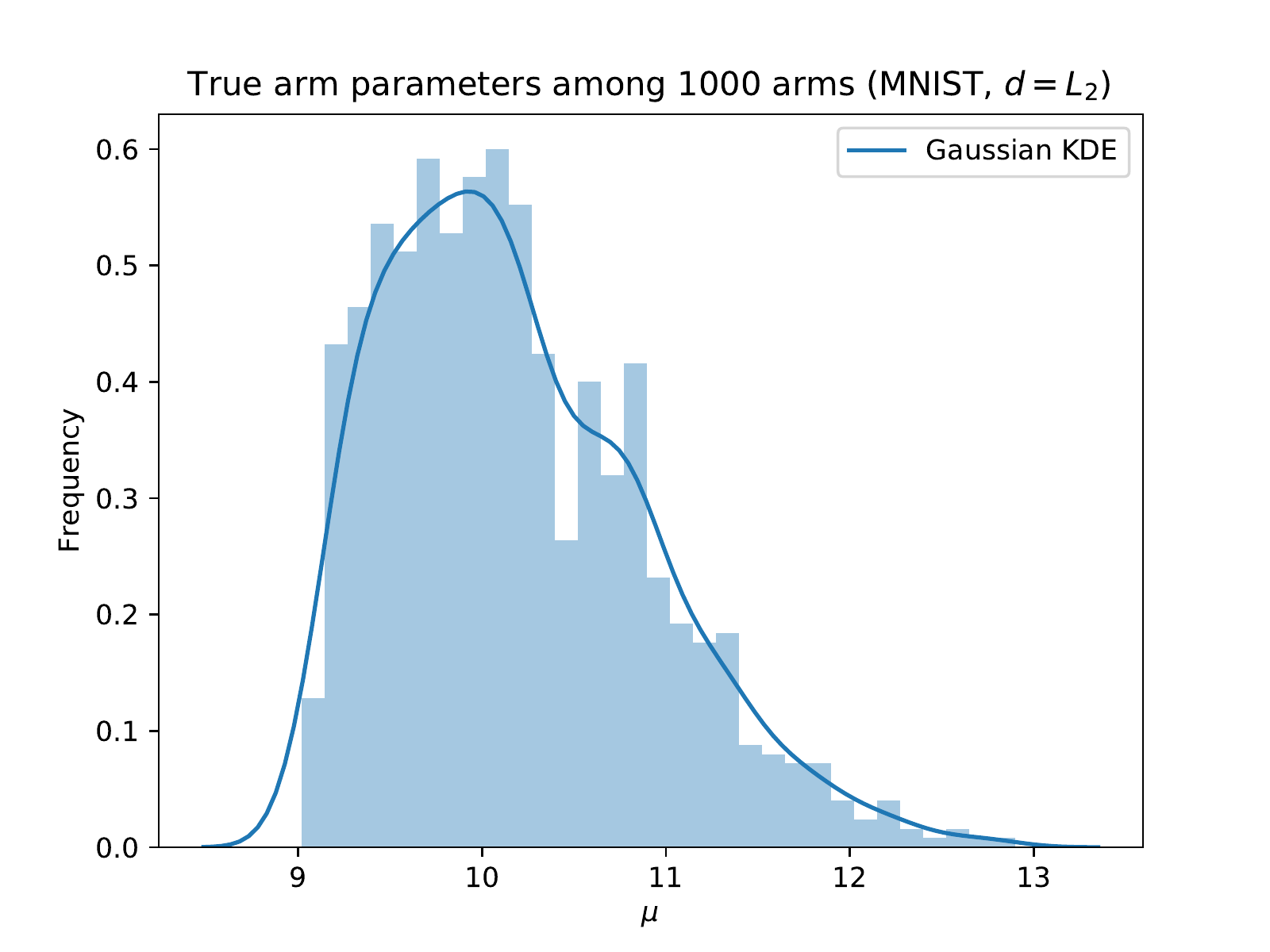}   
  \label{fig:mu_dist2}
\end{subfigure}
\begin{subfigure}{.5\textwidth}
  \centering
  \includegraphics[width=\linewidth]{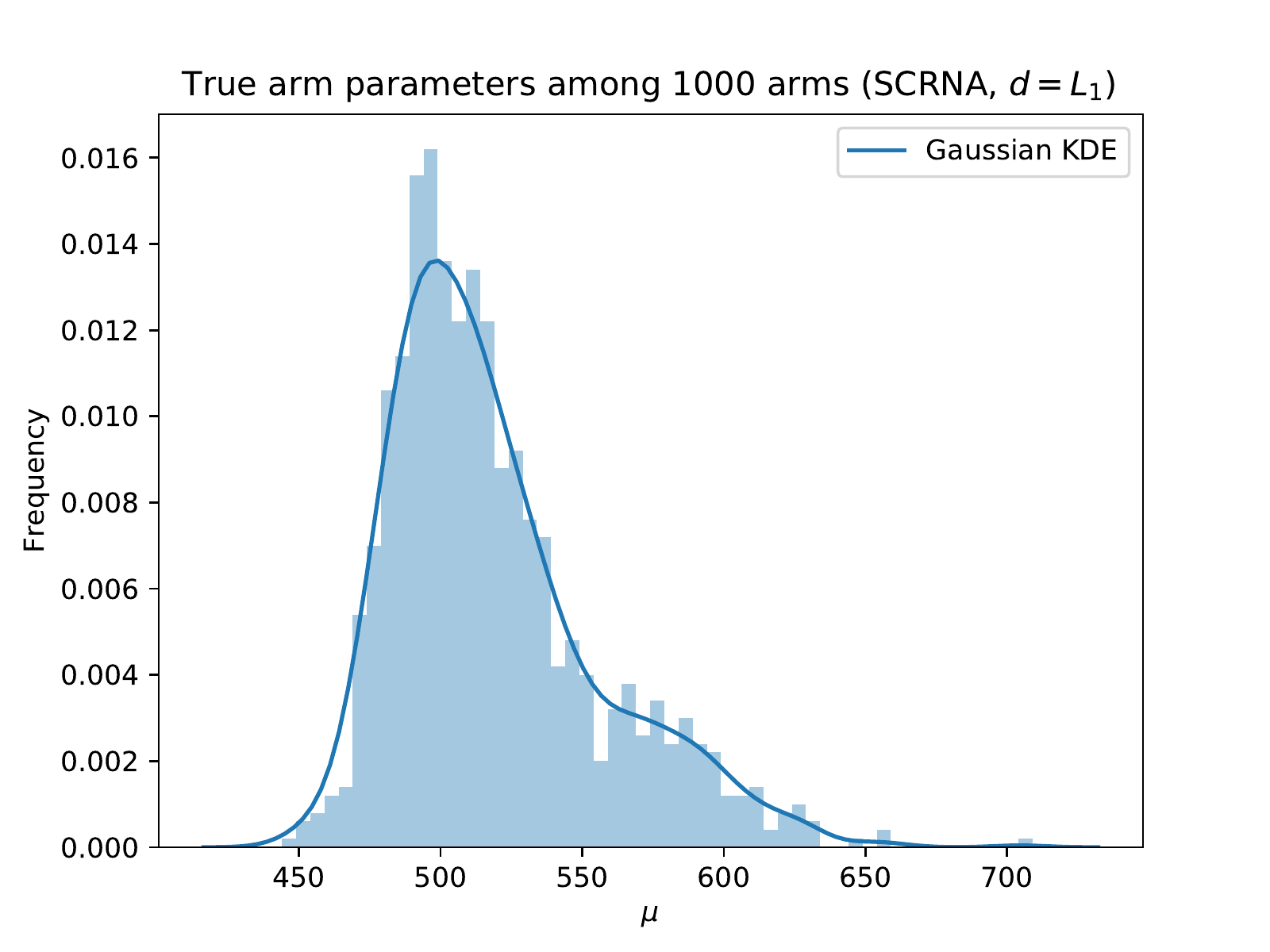}   
  \label{fig:mu_dist3}
\end{subfigure}
\begin{subfigure}{.5\textwidth}
  \centering
  \includegraphics[width=\linewidth]{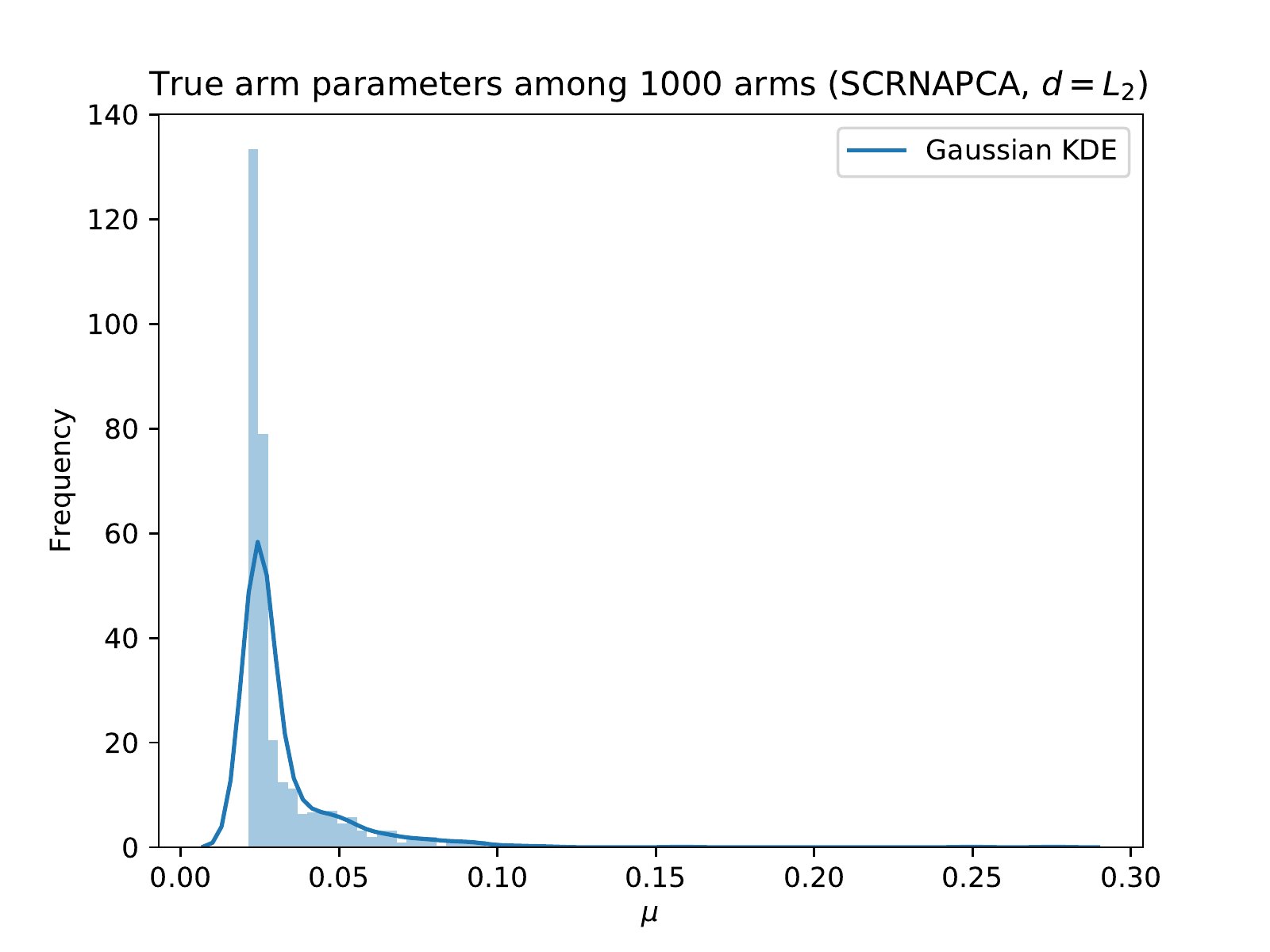}   
  \label{fig:mu_dist4}
\end{subfigure}
\caption{Histogram of true arm parameters, $\mu_i$, for 1000 randomly sampled arms in the first BUILD step of various datasets. For scRNA-PCA with $d = l_2$ (bottom right), the arm returns are much more sharply peaked about the mininum than for the other datasets. In plots where the bin widths are less than 1, the frequencies can be greater than 1.}
\label{fig:mu_dist}
\end{figure}

Figure \ref{fig:mu_dist} shows the distribution of arm parameters for various (dataset, metric) pairs in the first BUILD step. In this step, the arm parameter corresponds to the mean distance from the point (the arm) to every other point. We note that the true arm parameters in scRNA-PCA are more heavily concentrated about the minimum than in the other datasets. Intuitively, we have projected the points from a 10,170-dimensional space into a 10-dimensional one and have lost significant information in the process. This makes many points appear "similar" in the projected space.

\begin{figure}[ht]
\begin{subfigure}{.5\textwidth}
  \centering
  \includegraphics[width=\linewidth]{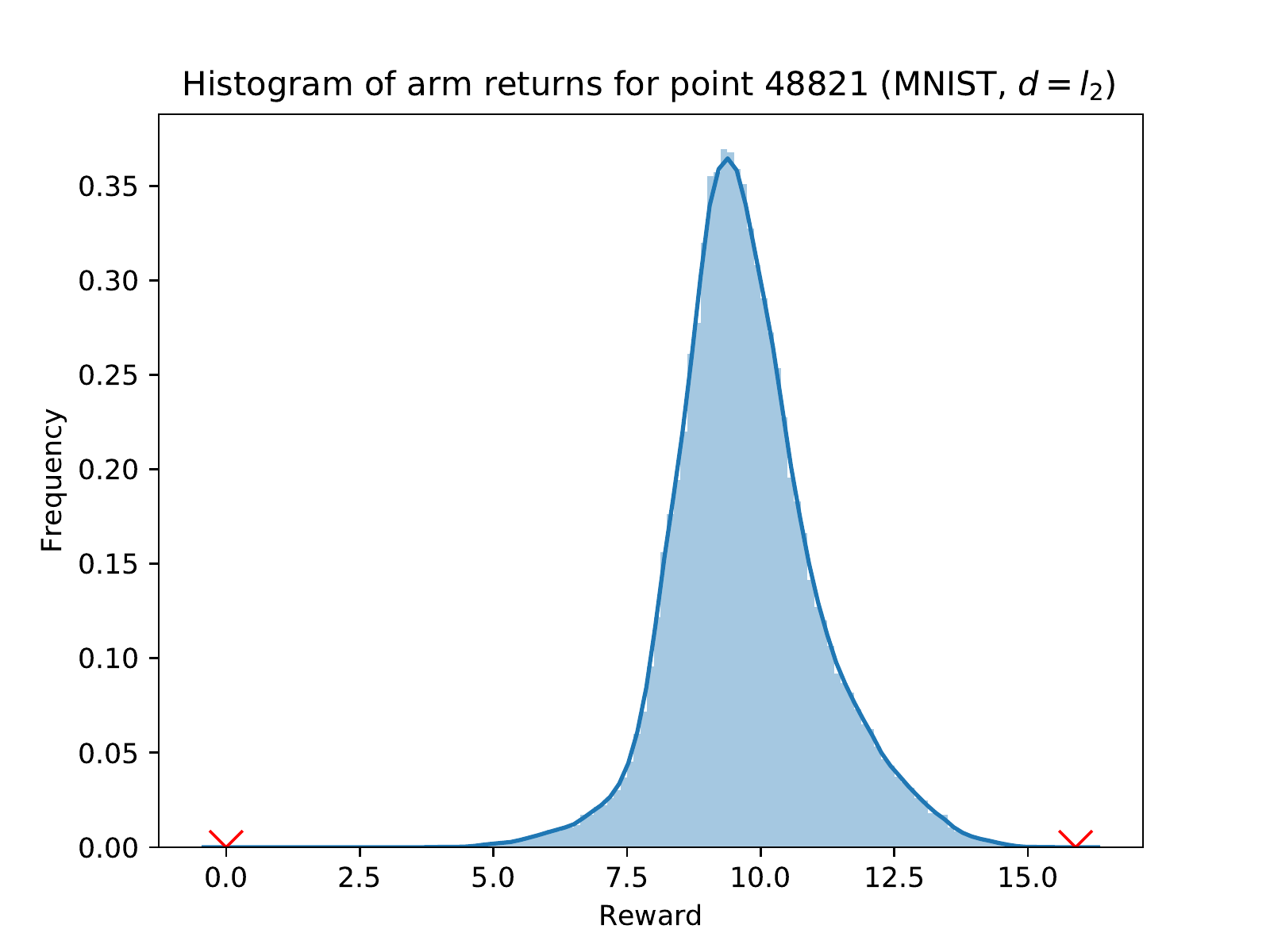}  
\end{subfigure}
\begin{subfigure}{.5\textwidth}
  \centering
  \includegraphics[width=\linewidth]{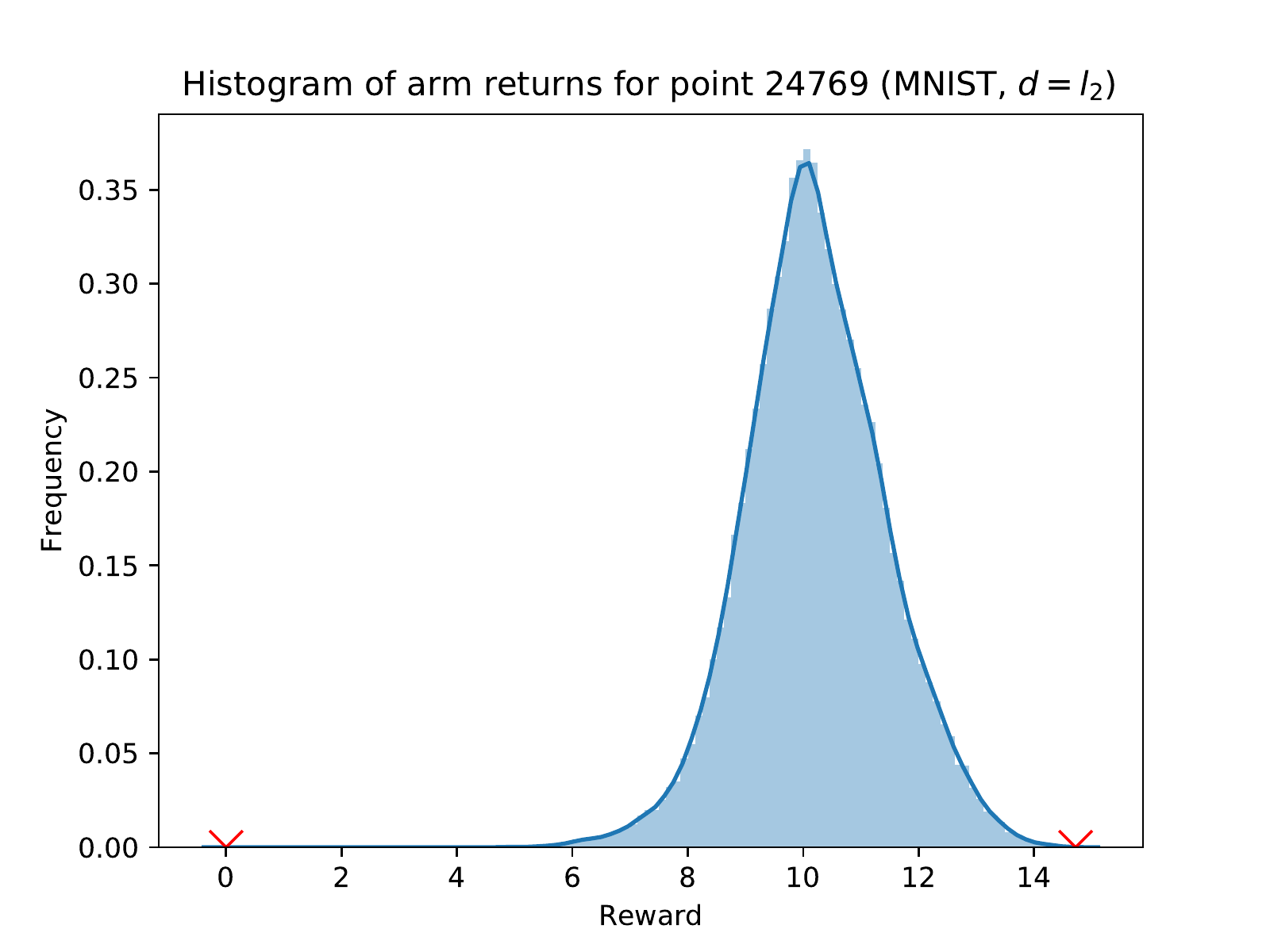}   
\end{subfigure}
\begin{subfigure}{.5\textwidth}
  \centering
  \includegraphics[width=\linewidth]{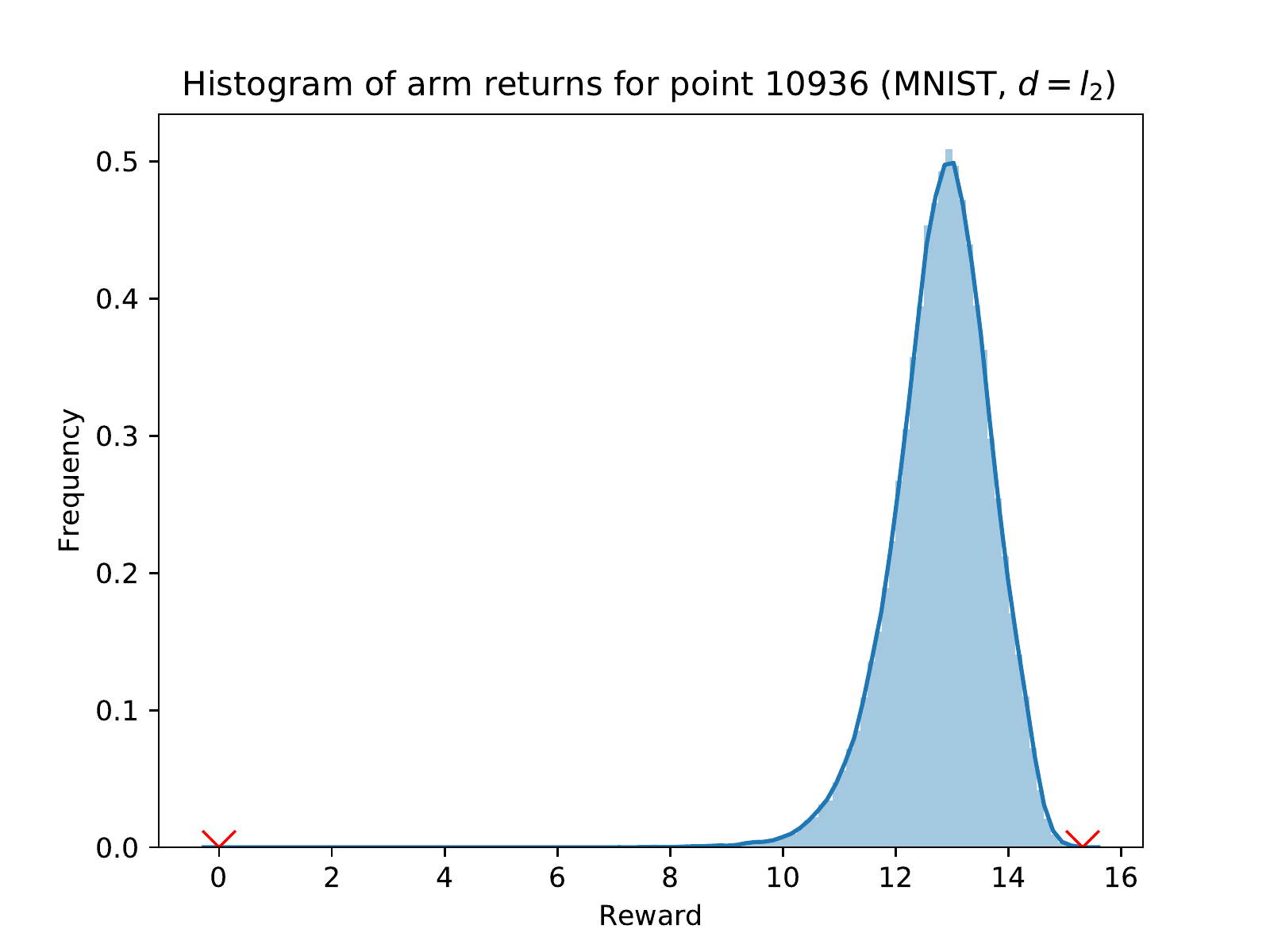}   
\end{subfigure}
\begin{subfigure}{.5\textwidth}
  \centering
  \includegraphics[width=\linewidth]{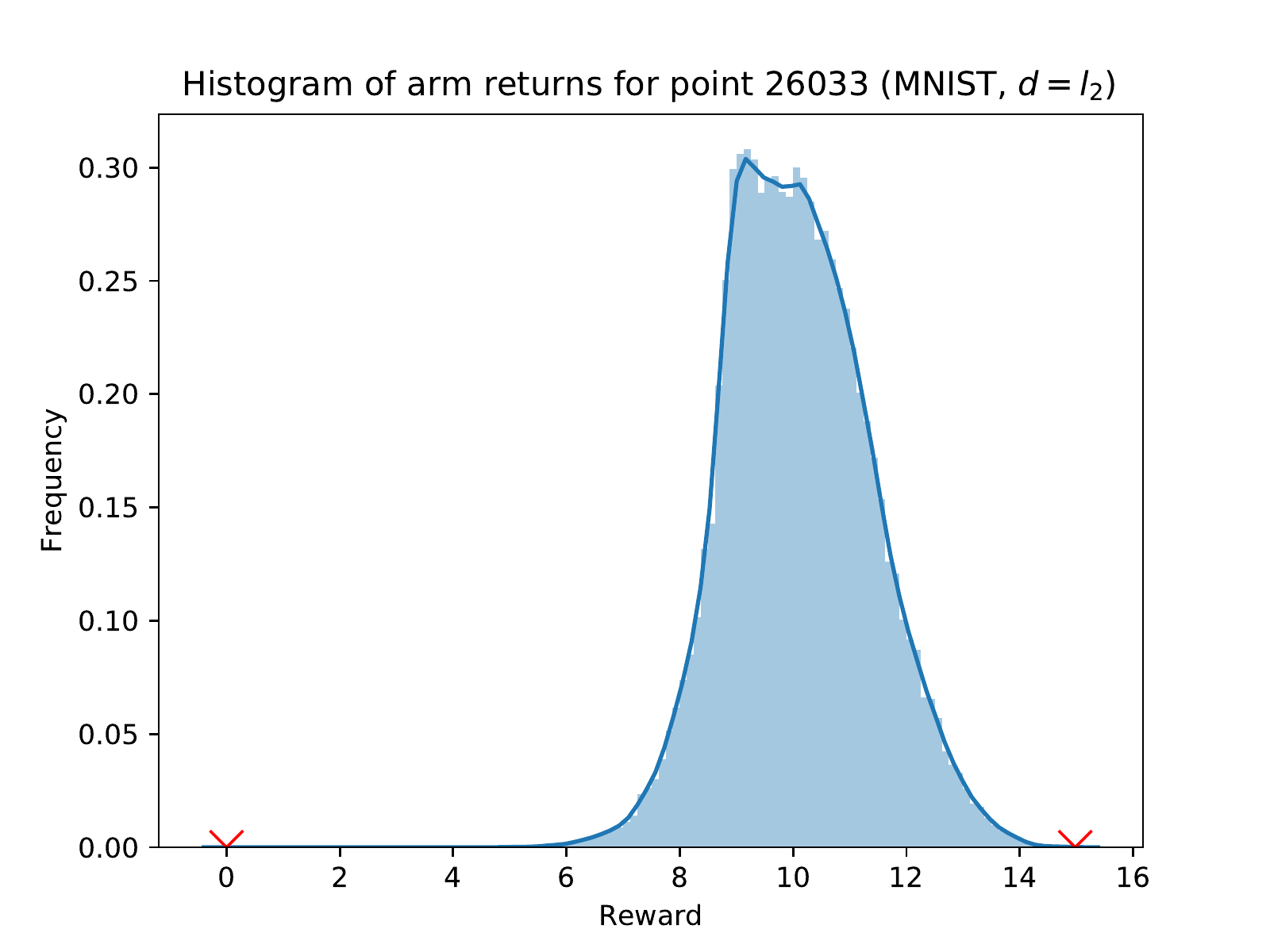}   
\end{subfigure}
\caption{Example distribution of rewards for 4 points in MNIST in the first BUILD step. The minimums and maximums are indicated with red markers.}
\label{fig:sigma_ex_MNIST}
\end{figure}

\begin{figure}[ht]
\begin{subfigure}{.5\textwidth}
  \centering
  \includegraphics[width=\linewidth]{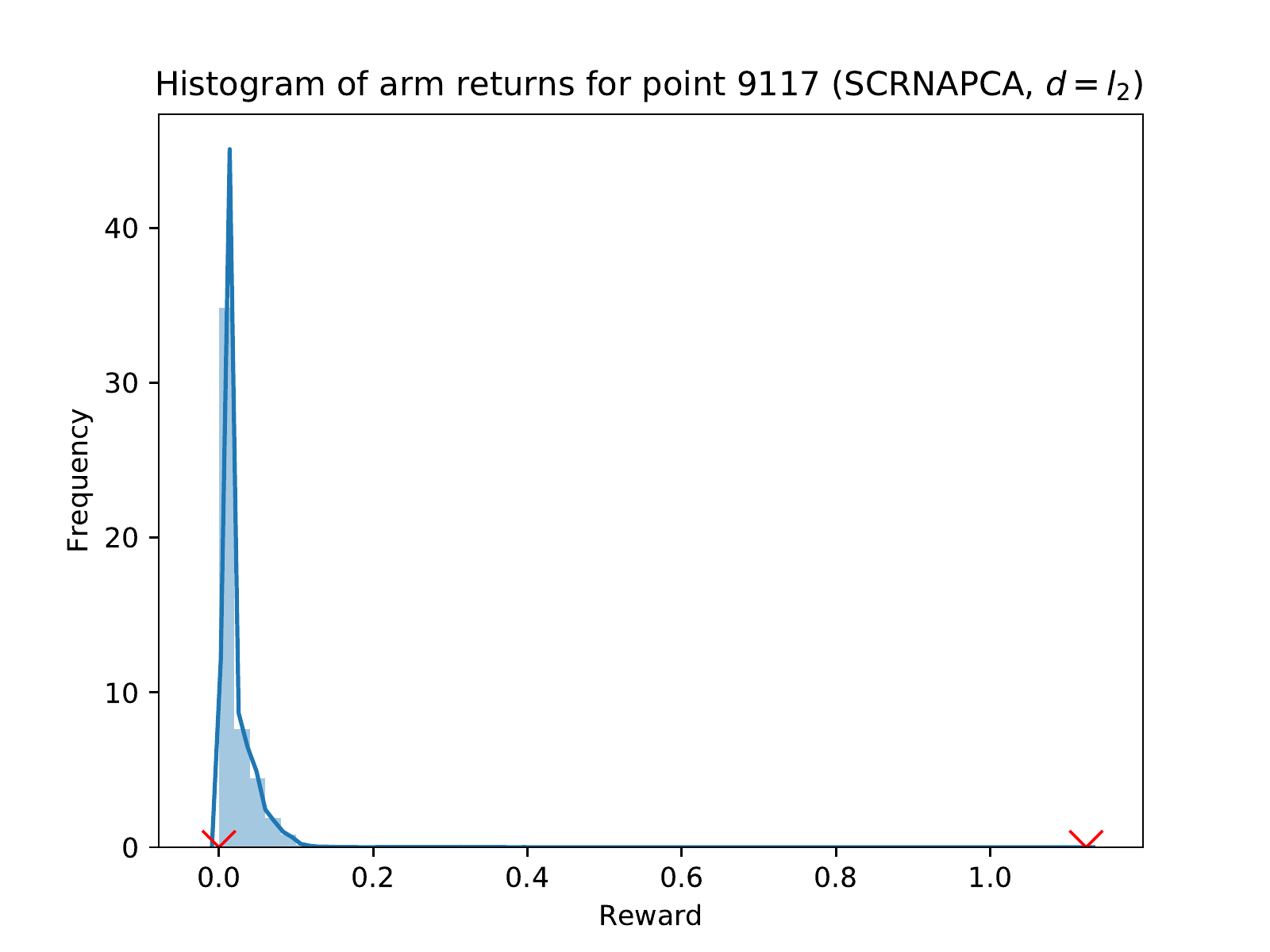}  
\end{subfigure}
\begin{subfigure}{.5\textwidth}
  \centering
  \includegraphics[width=\linewidth]{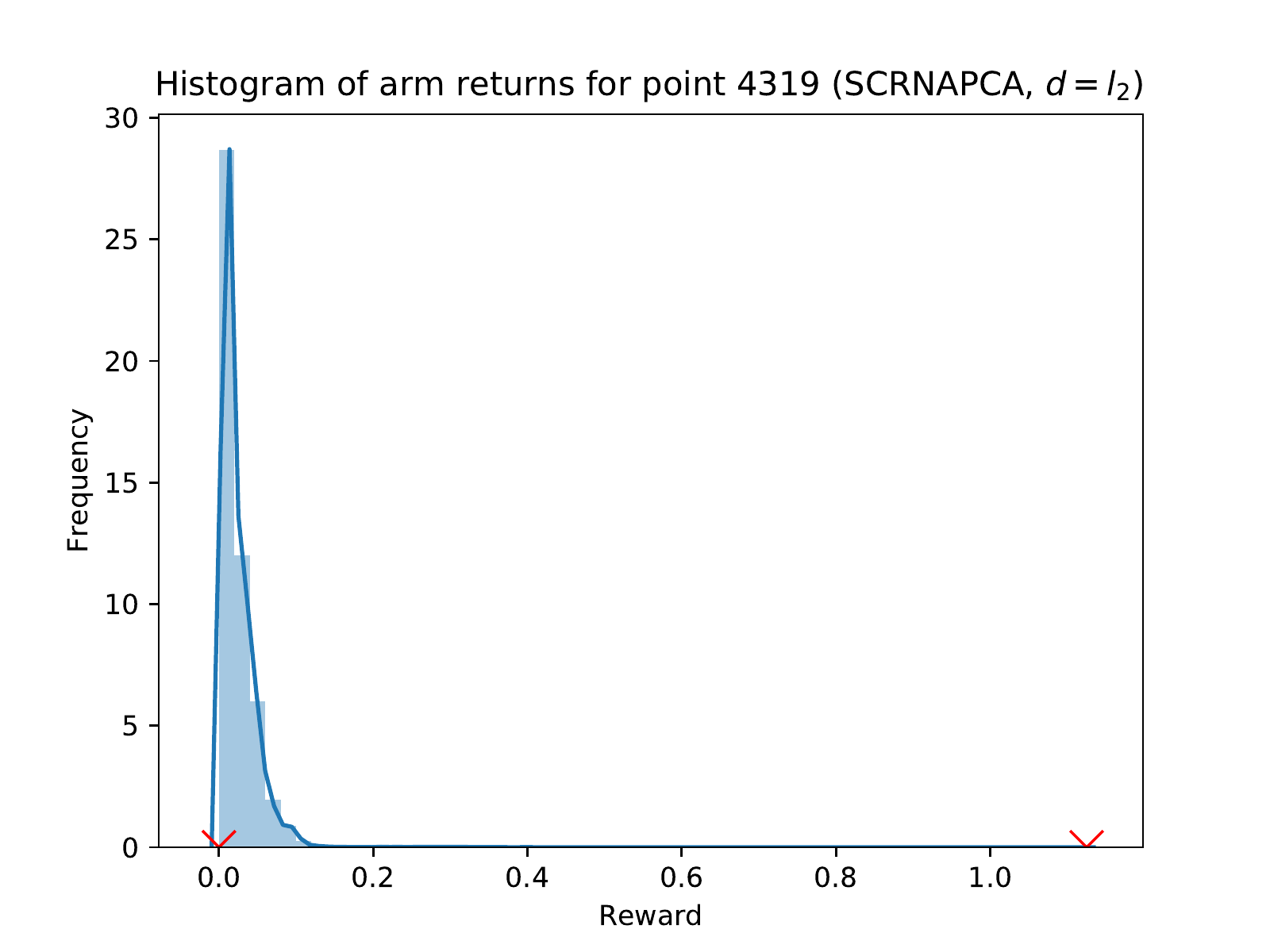}   
\end{subfigure}
\begin{subfigure}{.5\textwidth}
  \centering
  \includegraphics[width=\linewidth]{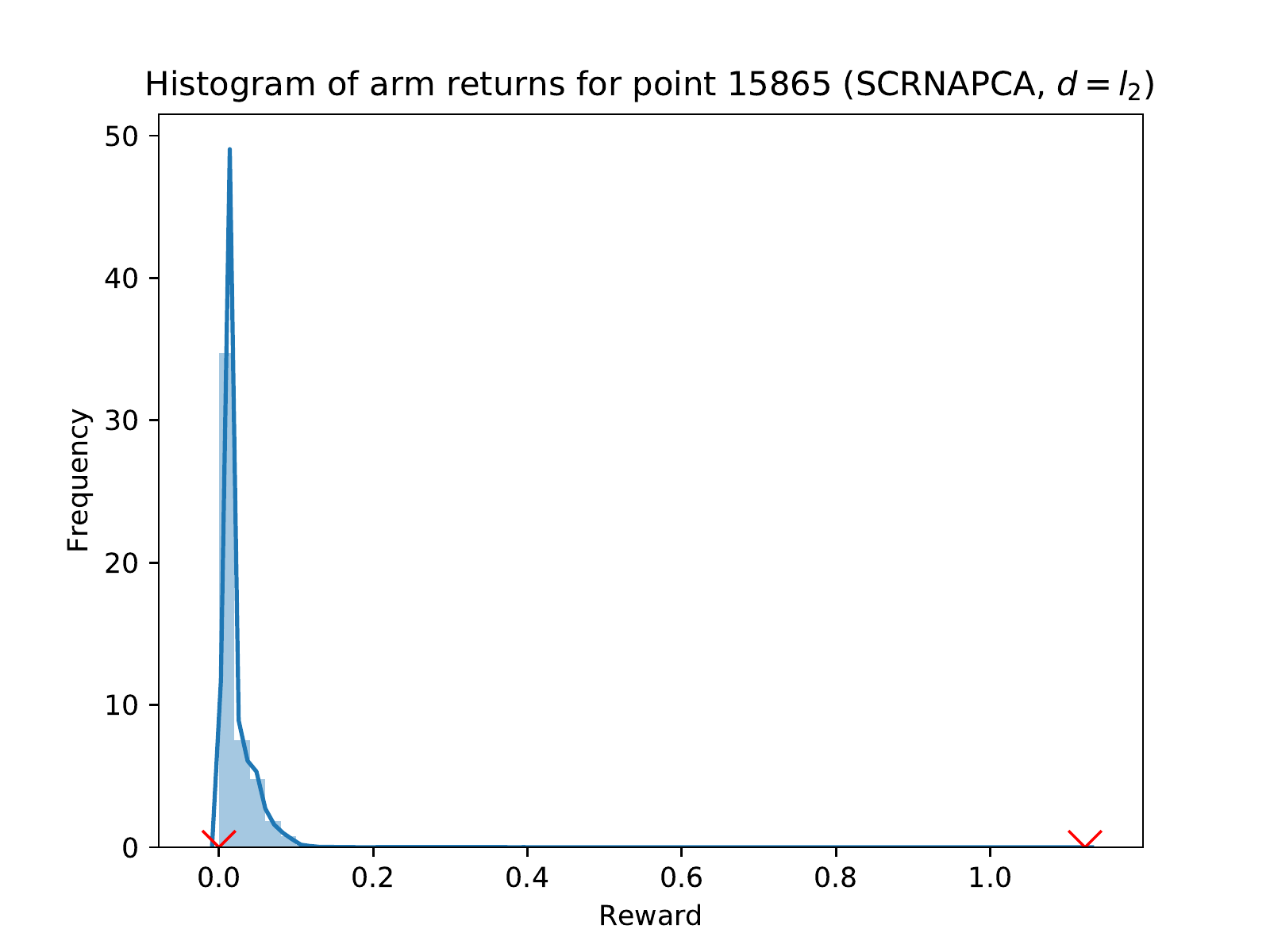}   
\end{subfigure}
\begin{subfigure}{.5\textwidth}
  \centering
  \includegraphics[width=\linewidth]{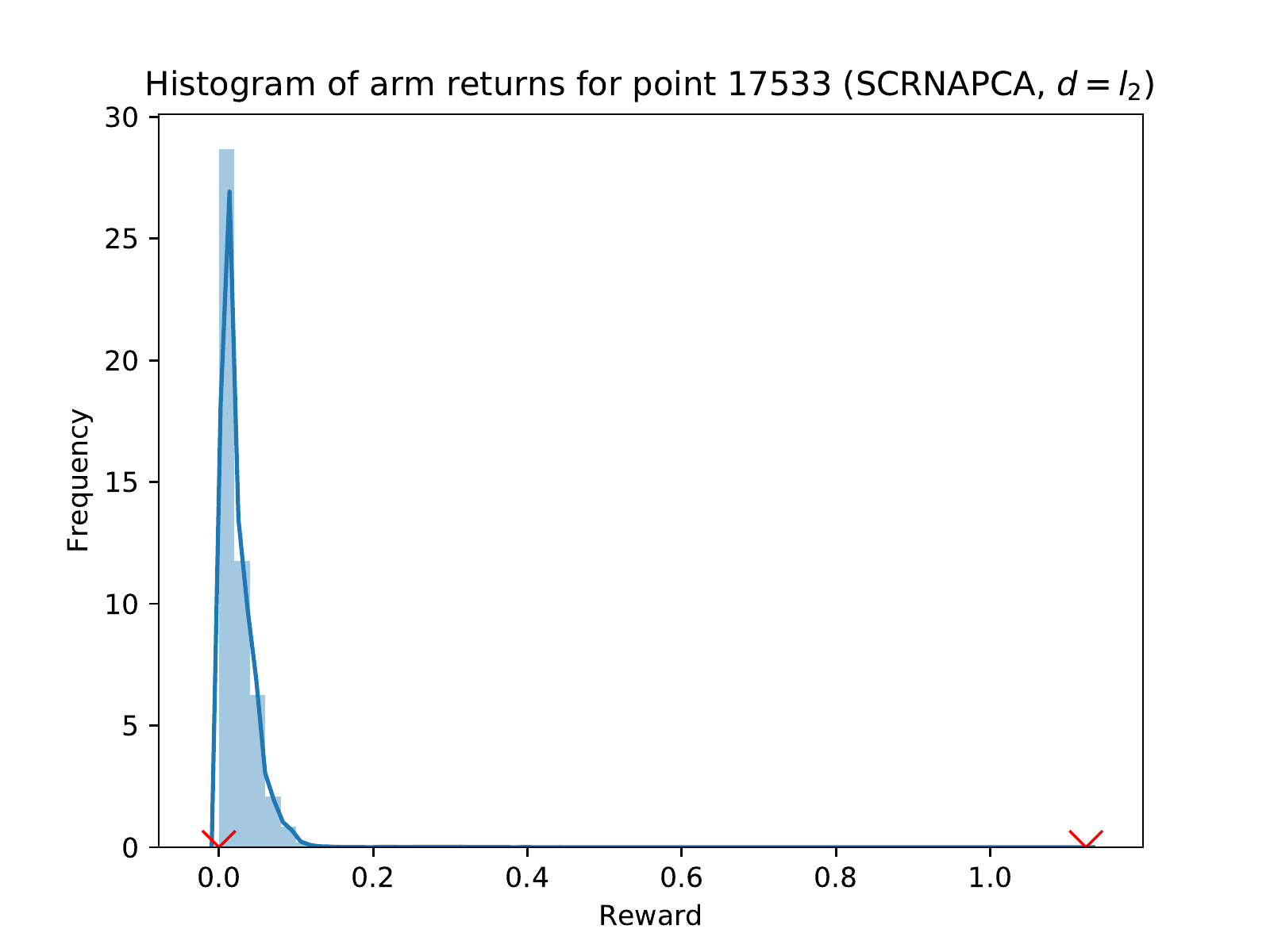}   
\end{subfigure}
\caption{Example distribution of rewards for 4 points in scRNA-PCA in the first BUILD step. The minimums and maximums are indicated with red markers. The distributions shown here are more heavy-tailed than in Figure \ref{fig:sigma_ex_MNIST}. In plots where the bin widths are less than 1, the frequencies can be greater than 1.}
\label{fig:sigma_ex_SCRNAPCA}
\end{figure}

Figures \ref{fig:sigma_ex_MNIST} and \ref{fig:sigma_ex_SCRNAPCA} show the distribution of arm rewards for 4 arms (points) in MNIST and scRNA-PCA, respectively, in the first BUILD step. We note that the examples from scRNA-PCA display much larger tails, suggesting that their sub-Gaussianity parameters $\sigma_x$ are very high.

\begin{figure}[ht!]
    \centering
    \includegraphics[scale=0.5]{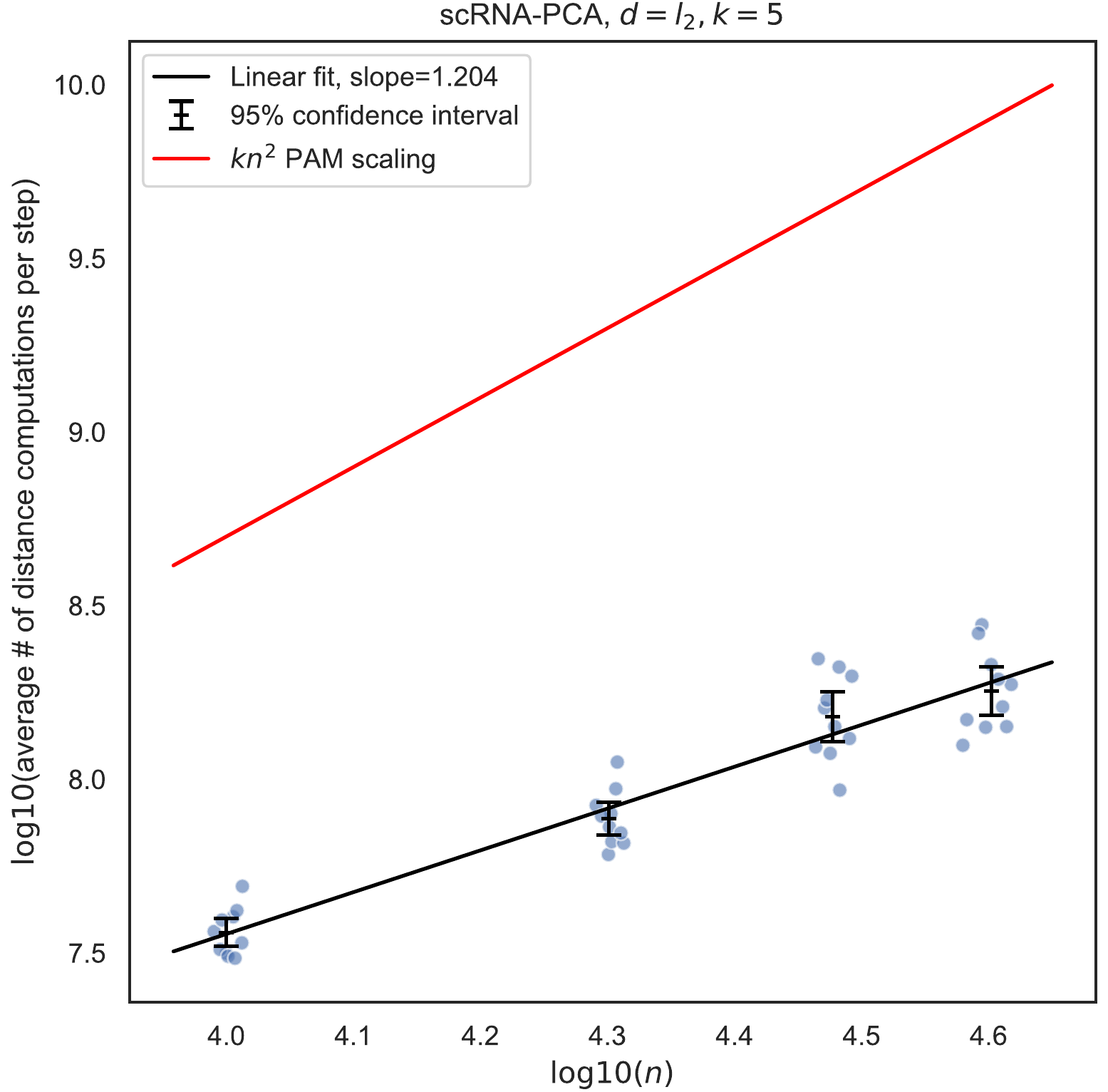}
    \caption{Average number of distance calls per iteration vs. $n$, for scRNA-PCA and $l_2$ distance on a log-log scale. The line of best fit (black) are plotted, as are reference lines demonstrating the expected scaling of PAM (red).} 
    \label{fig:SCRNAPCA-L2-scaling}
\end{figure}

Together, these observations suggest that the scRNA-PCA dataset may violate the assumptions of Theorems \ref{thm:specific} and \ref{thm:nlogn} and hurt the scaling of \algname with $n$, as measured by the number of distance calls per iteration as in Section \ref{sec:exps}. Figure \ref{fig:SCRNAPCA-L2-scaling} demonstrates the scaling of \algname with $n$ on scRNA-PCA. The slope of the line of best fit is 1.204, suggesting that \algname scales as approximately $O(n^{1.2})$ in dataset size. We note that this is higher than the exponents suggested for other datasets by Figures \ref{fig:mnist-l2} and \ref{fig:mnist-scrna}, likely to the different distributional characteristics of the arm means and their spreads.

We note that, in general, it may be possible to characterize the distribution of arm returns $\mu_i$ at and the distribution of $\sigma_x$, the sub-Gaussianity parameter, at every step of \algnamenospace, from properties of the data-generating distribution, as done for several distributions in \cite{bagaria2018medoids}. We leave this more general problem, as well as its implications for the complexity of our \algnamenospace, to future work.

\newpage
\clearpage
\section{Future Work}
\label{sec:future}



There are several ways in which \algname could be improved or made more impactful. In this work, we chose to implement a UCB-based algorithm to find the medoids of a dataset. Other best-arm-identification approaches, however, could also be used for this problem. 
It may also be possible to generalize a recent single-medoid approach, Correlation-Based Sequential Halving \cite{baharav2019ultra}, to more than $1$ medoid, especially to relax the sub-Gaussianity assumptions (discussed further in Appendix \ref{app:relaxation}). Though we do not have reason to suspect an algorithmic speedup (as measured by big-$O$), we may see constant factor or wall clock time improvements. We also note that it may be possible to prove the optimality of \algname in regards to algorithmic complexity, up to constant factors, using techniques from \cite{bagaria2018medoids} that were developed for sample-efficiency guarantees in hypothesis testing.



We also note that it may be possible to improve the theoretical bounds presented in Theorem 1; indeed, in experiments, a much larger error threshold $\delta$ was acceptable, which suggests that the bounds are weak; we discuss the hyperparameter $\delta$ further in Appendix \ref{appendix:banditpam-approx}. 

\subsection{Relaxing the sub-Gaussianity assumption \label{app:relaxation}}
An alternate approach using bootstrap-based bandits \cite{wang2020bootstrap, kveton2019abootstrap, kveton2019bbootstrap} could be valuable in relaxing the distributional assumptions on the data that the quantities of interest are $\sigma$-sub-Gaussian. Alternatively, if a bound on the distances is known, it may be possible to avoid the estimation of $\sigma_x$ by using the empirical Bernstein inequality to bound the number of distance computations per point, similar to how Hoeffding's inequality was used in the proof of Theorem \ref{thm:specific}. It may also be possible to use a related method from \cite{bestofbothworlds} to avoid these statistical assumptions entirely.

\subsection{Intelligent Cache Design \label{appendix:cache}}
The existing implementation does not cache pairwise distance computations, despite the fact that \algname spends upwards of 98\% of its runtime in evaluating distances, particularly when such distances are expensive to compute. This is in stark contrast to the state-of-the-art implementations of PAM, FastPAM1, and FastPAM1, which precompute and cache the entire $n^2$ distance matrix before any medoid assignments are made.

It should be possible to implement a cache in \algname that would dramatically reduce wall-clock-time requirements. Furthermore, it may be possible to cache only $O(n\log n)$ pairwise distances, instead of all $n^2$ distances. This could be done, for example, by fixing an ordering of the reference points to be used in each call to Algorithm \ref{alg:bandit_based_search}. Since, on average, only $O(\log n)$ reference points are required for each target point, it should not be necessary to cache all $n^2$ pairwise distances. Furthermore, the same cache could be used across different calls to Algorithm \ref{alg:bandit_based_search}, particularly since we did not require independence of the sampling of the reference points in the proof of Theorem \ref{thm:nlogn}. Finally, it may be possible to reduce this cache size further by using techniques from \cite{newlingSeeding}.

\subsection{Approximate version of \algname}
\label{appendix:banditpam-approx}
We note that \algname is a randomized algorithm which requires the specification of the hyperparameter $\delta$. Intuitively, $\delta$ governs the error probability that \algname returns a suboptimal target point $x$ in any call to Algorithm \ref{alg:bandit_based_search}. The error parameter $\delta$ suggests the possibility for an approximate version of \algname that may not required to return the same results as PAM. If some concessions in final clustering loss are acceptable, $\delta$ can be increased to improve the runtime of \algnamenospace. An analysis of the tradeoff between the final clustering loss and runtime of \algnamenospace, governed by $\delta$, is left to future work. It may also be possible to combine the techniques in \cite{pamlite} with \algname to develop an approximate algorithm.

\subsection{Dependence on $d$}
\label{appendix:scalingwithd}
Throughout this work, we assumed that computing the distance between two points was an $O(1)$ operation. This obfuscates the dependence on the dimensionality of the data, $d$. If we consider computing the distance between two points an $O(d)$ computation, the complexity of \algname could be expressed as $O(dn$log$n)$ in the BUILD step and each SWAP iteration. Recent work \cite{bagaria2018adaptive} suggests that this could be further improved; instead of computing the difference in each of the $d$ coordinates, we may be able to adaptively sample which of the $d$ coordinates to use in our distance computations and reduce the dependence on dimensionality from $d$ to $O(\log d)$, especially in the case of sparse data.

\subsection{Dependence on $k$}
\label{appendix:scalingwithk}

In this paper, we treated $k$ as a constant in our analysis and did not analyze the explicit dependence of \algname on $k$. In experiments, we observed that the runtime of \algname scaled linearly in $k$ in each call to Algorithm \ref{alg:bandit_based_search} when $k \ll n$. We also observe \algname scales linearly with $k$ when $k$ is less than the number of "natural" clusters of the dataset (e.g. $\sim$10 for MNIST). However, we were able to find other parameter regimes where the scaling of Algorithm \ref{alg:bandit_based_search} with $k$ appears quadratic. Furthermore, we generally observed that the number of swap steps $T$ required for convergence was $O(k)$, consistent with \cite{schubert2019faster}, which could make the overall scaling of \algname with $k$ superlinear when each call to Algorithm \ref{alg:bandit_based_search} is also $O(k)$. We emphasize that these are only empirical observations. We leave a formal analysis of the dependence of the overall \algname algorithm on $k$ to future work.

\clearpage
\section{Proofs of Theorems~\ref{thm:specific} and \ref{thm:nlogn}}
\label{app:thmproof}
\setcounter{theorem}{0}

\begin{theorem} \label{thm:specific}
For $\delta = n^{-3}$, with probability at least $1-\tfrac{2}{n}$, Algorithm \ref{alg:bandit_based_search}
returns the correct solution to \eqref{eqn:build_search} (for a BUILD step) or \eqref{eqn:swap_search} (for a SWAP step),
using a total of $M$ distance computations, where
\aln{
E[M] \leq 4n + \sum_{x \in \X}  \min \left[ \frac{12}{\Delta_x^2} \left(\sigma_x+\sigma_{x^*} \right)^2 \log n + B, 2n \right].
}
\end{theorem}


\begin{proof}
First, we show that, with probability at least $1-\tfrac{2}{n}$, all confidence intervals computed throughout the algorithm are true confidence intervals, in the sense that they contain the true parameter $\mu_x$.
To see this, notice that for a fixed $x$ and a fixed iteration of the algorithm, $\hat \mu_x$ is the average of $\nuref$ i.i.d.~samples of a $\sigma_x$-sub-Gaussian distribution.
From Hoeffding's inequality, 
\aln{
\Pr\left( \left| \mu_x - \hat \mu_x \right| > C_x \right) \leq 2 \exp \left({-\frac{\nuref C_x^2}{2\sigma_x^2}}\right)  =\vcentcolon 2 \delta.
}
Note that there are at most $\frac{n^2}{B} \leq n^2$ such confidence intervals computed across all target points (i.e. arms) and all steps of the algorithm, where $B$ is the batch size.
If we set $\delta = 1/n^3$, we see that $\mu_x \in [\hat \mu_x - C_x, \hat \mu_x + C_x]$ for every $x$ and for every step of the algorithm with probability at least $1-\frac{2}{n}$, by the union bound over at most $n^2$ confidence intervals.

Next, we prove the correctness of Algorithm \ref{alg:bandit_based_search}.
Let $x^* = \argmin_{x \in \Star} \mu_x$ be the desired output of the algorithm.
First, observe that the main \texttt{while} loop in the algorithm can only run $\frac{n}{B}$ times, so the algorithm must terminate.
Furthermore, if all confidence intervals throughout the algorithm are correct, it is impossible for $x^*$ to be removed from the set of candidate target points. 
Hence, $x^*$ (or some $y \in \Star$ with $\mu_y = \mu_{x^*}$) must be returned upon termination with probability at least $1-\frac{2}{n}$.

Finally, we consider the complexity of Algorithm \ref{alg:bandit_based_search}. 
Let $\nuref$ be the total number of arm pulls computed for each of the arms remaining in the set of candidate arms at some point in the algorithm.
Notice that, for any suboptimal arm $x \ne x^*$ that has not left the set of candidate arms, we must have
$C_x = \sigma_x \sqrt{ \log(\tfrac{1}{\delta}) /\nuref}$.
With $\delta = n^{-3}$ as above and $\Delta_x \vcentcolon= \mu_x - \mu_{x^*}$, if $\nuref > \frac{12}{\Delta_x^2} \left(\sigma_x+\sigma_{x^*}\right)^2 \log n$,
then
\aln{
2(C_x + C_{x^*}) = 2 \left( \sigma_x + \sigma_{x^*}\right) \sqrt{  { \log(n^3) } / {\nuref  }} < \Delta_x = \mu_x - \mu_{x^*},
}
and
\begin{align*}
    \hat \mu_x - C_x &> \mu_x - 2C_x \\
    &= \mu_{x^*} + \Delta_x - 2C_x \\
    &\geq \mu_{x^*} + 2 C_{x^*} \\
    &> \hat \mu_{x^*} + C_{x^*}
\end{align*}


implying that $x$ must be removed from the set of candidate arms at the end of that iteration.
Hence, the number of distance computations $M_x$ required for target point $x \ne x^*$ is at most
\aln{
M_x \leq \min \left[ \frac{12}{\Delta_x^2} \left( \sigma_x + \sigma_{x^*}\right)^2 \log n + B, 2n \right].
}
Notice that this holds simultaneously for all $x \in \Star$ with probability at least $1-\tfrac{2}{n}$.
We conclude that the total number of distance computations $M$ satisfies
\aln{
E[M] & \leq E[M | \text{ all confidence intervals are correct}] + \frac{2}{n} (2n^2) \\
& \leq 4n + \sum_{x \in \X}  \min \left[ \frac{12}{\Delta_x^2} \left( \sigma_x + \sigma_{x^*}\right)^2 \log n + B, 2n \right]
}
where we used the fact that the maximum number of distance computations per target point is $2n$.
\end{proof}

\textbf{Remark A1:} An analogous claim can be made for arbitrary $\delta$. For arbitrary $\delta$, the probability that all confidence intervals are true confidence intervals is at least $1 - 2n^2 \delta$, and the expression for $E[M]$ becomes:
\aln{
E[M] & \leq E[M | \text{ all confidence intervals are correct}] + 4n^4 \delta \\
& \leq 4n^4 \delta + \sum_{x \in \X}  \min \left[ \frac{4}{\Delta_x^2} \left( \sigma_x + \sigma_{x^*}\right)^2 \log (\frac{1}{\delta}) + B, 2n \right]
}

\begin{theorem} \label{thm:nlogn}
If \algname is run on a dataset $\X$ with $\delta = n^{-3}$, then it returns the same set of $k$ medoids as PAM with probability $1-o(1)$. 
Furthermore, 
the total number of distance computations $M_{\rm total}$ required satisfies
\aln{
E[M_{\rm total}] = O\left( n \log n \right).
}
\end{theorem}

From Theorem \ref{thm:specific}, the probability that Algorithm \ref{alg:bandit_based_search} does not return the target point $x$ with the smallest value of $\mu_x$ in a single call, i.e. that the result of Algorithm \ref{alg:bandit_based_search} will differ from the corresponding step in PAM, is at most $2/n$.
By the union bound over all $k+T$ calls to Algorithm \ref{alg:bandit_based_search}, the probability that \algname does not return the same set of $k$ medoids as PAM is at most $2(k+T)/n = o(1)$, since $k$ and $T$ are taken as constants. This proves the first claim of Theorem \ref{thm:nlogn}.

It remains to show that $E[M_{\rm total}] = O( n \log n)$. Note that, if a random variable is $\sigma$-sub-Gaussian, it is also $\sigma'$-sub-Gaussian for $\sigma' > \sigma$.
Hence, if we have a universal upper bound $\sigma_{\rm ub}> \sigma_x$ for all $x$, Algorithm \ref{alg:bandit_based_search} can be run with $\sigma_{\rm ub}$ replacing each $\sigma_x$.
In that case, a direct consequence of Theorem \ref{thm:specific} is that the total number of distance computations per call to Algorithm \ref{alg:bandit_based_search} satisfies
\al{
E[M] & \leq 4n + \sum_{x \in \X}  48 \frac{\sigma^2_{\rm ub}}{\Delta_x^2} \log n + B 
\leq  4n + 48 \left(\frac{\sigma_{\rm ub}}{\min_x \Delta_x}\right)^2
n \log n.
\label{eq:expectedM}
}
Furthermore, as proven in Appendix 2 of \cite{bagaria2018medoids}, such an instance-wise bound, which depends on the $\Delta_x$s, converts to an $O(n \log n)$ bound when the $\mu_x$s follow a sub-Gaussian distribution. 
Moreover, since at most $k+T$ calls to Algorithm \ref{alg:bandit_based_search} are made, from \eqref{eq:expectedM} we see that the total number of distance computations $M_{\rm total}$ required by \algname satisfies $E[M_{\rm total}] = O( n \log n)$.

\end{document}